\documentclass[10pt,twocolumn,letterpaper]{article}

\usepackage[pagenumbers]{cvpr} 

\usepackage[dvipsnames]{xcolor}

\usepackage{amsmath}
\usepackage{dsfont}

\usepackage{multirow}
\usepackage{pifont}
\usepackage{color,colortbl}
\definecolor{Gray}{gray}{0.90}

\definecolor{cvprblue}{rgb}{0.21,0.49,0.74}
\usepackage[pagebackref,breaklinks,colorlinks,allcolors=cvprblue]{hyperref}

\def\paperID{8918} 
\def\confName{CVPR}
\def\confYear{2025}

\title{Steady Progress Beats Stagnation: Mutual Aid of Foundation and Conventional Models in Mixed Domain Semi-Supervised Medical Image Segmentation}

\author{
Qinghe Ma$^1$,~~Jian Zhang$^1$,~~Zekun Li$^1$,~~Lei Qi$^2$,~~Qian Yu$^3$,~~Yinghuan Shi$^{1,}$\thanks{Corresponding author: Yinghuan Shi (syh@nju.edu.cn). Qinghe Ma, Jian Zhang, Zekun Li and Yinghuan Shi are with the State Key Laboratory for Novel Software Technology and National Institute of Healthcare Data Science, Nanjing University, China. 
}~\\
$^1$Nanjing University~~~~~$^2$Southeast University~~~~~$^3$Shandong Women's University\\
}

\begin{document}
\maketitle
\begin{abstract}
Large pretrained visual foundation models exhibit impressive general capabilities. However, the extensive prior knowledge inherent in these models can sometimes be a double-edged sword when adapting them to downstream tasks in specific domains.
In the context of semi-supervised medical image segmentation with domain shift, foundation models like MedSAM tend to make overconfident predictions, some of which are incorrect. The error accumulation hinders the effective utilization of unlabeled data and limits further improvements.
In this paper, we introduce a \underline{Syn}ergistic training framework for \underline{Fo}undation and \underline{C}onventional models (\textbf{SynFoC}) to address the issue. We observe that a conventional model trained from scratch has the ability to correct the high-confidence mispredictions of the foundation model, while the foundation model can supervise it with high-quality pseudo-labels in the early training stages. Furthermore, to enhance the collaborative training effectiveness of both models and promote reliable convergence towards optimization, the consensus-divergence consistency regularization is proposed. We demonstrate the superiority of our method across four public multi-domain datasets. In particular, our method improves the Dice score by 10.31\% on the Prostate dataset. Our code is available at \textcolor{magenta}{\url{https://github.com/MQinghe/SynFoC}}.
\vspace{-10pt}
\end{abstract}    
\section{Introduction}
\label{sec:intro}

\begin{figure}
    \centering
    \includegraphics[width=0.85\linewidth]{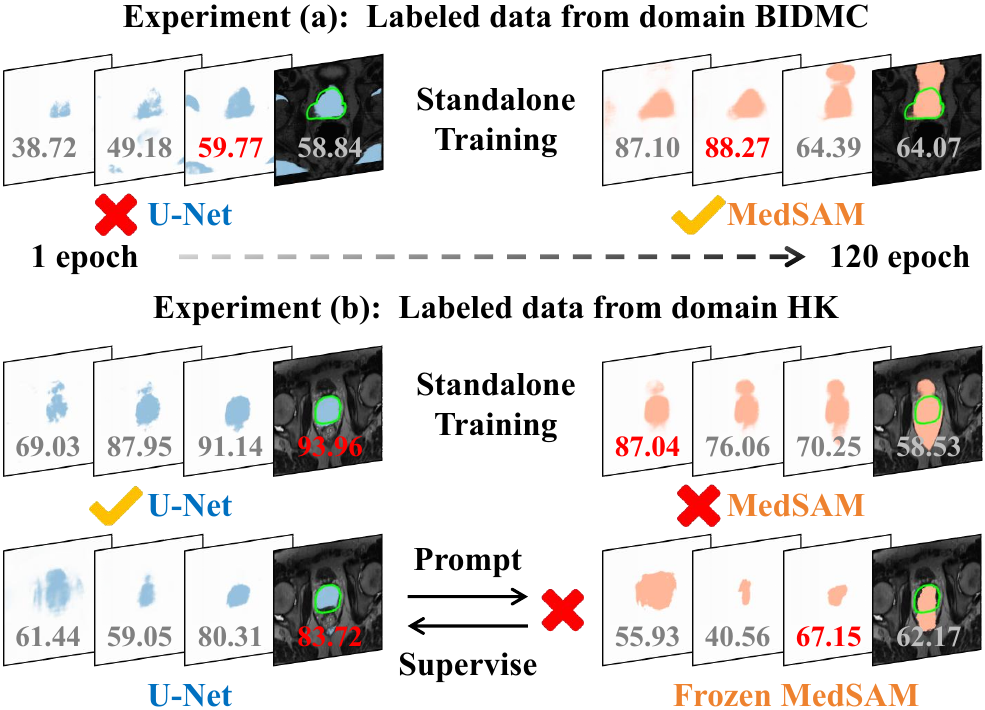}
    \vspace{-5pt}
    \caption{Illustration of pseudo-labels generation across training stages for various methods. In experiment (a) and (b) on prostate dataset, 20 labeled data come from BIDMC and HK, respectively. We report the Dice Coefficient between each pseudo-label and the ground truth, which is represented by the green contour. Standalone training of U-Net and MedSAM, as well as the guidance from MedSAM to U-Net, fail to effectively address MiDSS.}
    \label{fig:motivation}
    \vspace{-15pt}
\end{figure}


Semi-supervised medical image segmentation (SSMIS)~\cite{bai2017semi,yu2019uncertainty,fan2020inf,wu2022mutual,bai2023bidirectional,you2024rethinking,chi2024adaptive,qiu2024devil} offers an effective way to tackle problems with limited annotations~\cite {grunberg2017annotating,taleb20203d}. In recent years, domain shifts in real-world clinical scenarios have garnered significant attention, where labeled and unlabeled data are drawn from different distributions due to variations in equipment parameters, patient populations, and disease severity~\cite{zhang2022shifting,guo2022evaluation,yang2024limits}. 
When large amounts of medical data are collected, it is challenging to exam the distribution to which they belong. 
Consequently, many related settings have been explored~\cite{chen2020unsupervised, xie2022unsupervised, shin2023sdc,luo2017label,zhao2022uda,ma2024constructing}, with Mixed Domain Semi-Supervised Medical Image Segmentation (MiDSS) being a more general framework~\cite{ma2024constructing}.
In this setting, a limited number of labeled samples are sourced from a single domain, while a substantial number of unlabeled samples originate from multiple mixed domains.
For SSMIS, existing methods usually train a conventional model (\eg, U-Net~\cite{ronneberger2015u}) using labeled and unlabeled data from scratch. However, the domain shift 
renders the training of conventional models particularly vulnerable to noisy pseudo-labels~\cite{chen2022deliberated,ma2024constructing}. Recently, large pretrained visual foundation models~\cite{wang2023seggpt,khani2023slime,zou2024segment, ma2024segment} have demonstrated impressive segmentation performance and generalization capabilities for downstream tasks. Considering the limitations of conventional models, we wonder whether foundation models can serve as off-the-shelf tools for addressing these problems. In other words, can they be effectively adapted to specific domains by leveraging a small amount of labeled data alongside mixed-domain unlabeled data?

To answer the question, we conduct experiments to explore how does the single foundation model or conventional model perform in MiDSS. MedSAM~\cite{ma2024segment} is served as the foundation model and U-Net as the conventional model.
As illustrated in~\cref{fig:motivation}(a), when labeled data are drawn from the domain BIDMC, U-Net overfits to the labeled data, resulting in poor performance. In comparison, MedSAM exhibits superior segmentation capability in the early stages of training due to its inherent extensive prior knowledge. 
Similarly, as shown in~\cref{fig:motivation}(b), when labeled data comes from the domain HK, it fails to rectify high-confidence wrong predictions, hindering further performance improvement. In contrast, U-Net actively correct high-uncertainty mispredictions, achieving a higher performance ceiling.
\cref{fig:motivation}(a) and~\cref{fig:motivation}(b) show that neither the conventional nor the foundation model is universally effective. The conventional model tends to overfit to the labeled data when there are significant domain shifts between labeled and unlabeled data, while the foundation model, not limited to MedSAM struggles to correct high-confidence mispredictions due to large-scale pretraining, leading to error accumulation.
Additionally, many existing studies train a conventional model guided by the pseudo-labels from frozen foundation model. As shown in~\cref{fig:motivation}(b), pseudo-labels from U-Net provide bounding box prompts to MedSAM, which, in turn, offers additional supervisory signals for U-Net. However, under this training scheme, the performance of the conventional model is often limited by the foundation model.

Unlike previous studies where foundation models dominate~\cite{zhang2023semisam,zhang2023samdsk,chen2023aslseg,wang2025weakmedsam}, considering the complementary characteristics of both model, we believe that conventional models also play a critical role in further boosting the performance of foundation models.
In this paper, we propose a mixed-domain semi-supervised medical image segmentation \textbf{Syn}ergistic training framework where \textbf{Fo}undation (\eg, MedSAM) and \textbf{C}onventional (\eg, U-Net) models are synergistically trained (\textbf{SynFoC}). We dynamically adjusts the dominance of each model during training: MedSAM leads in the early stages to ensure training quality of U-Net, while U-Net takes the lead in later stages to correct high-confidence errors, unlocking performance potential of MedSAM.
Beyond that, to boost their representational abilities jointly, we employ consistency regularization~\cite{yu2019uncertainty,li2020transformation} to enhances information sharing. Yet, for regions with consistent predictions, encouraging higher confidence offers a more reliable optimization direction. Thus, we propose region-specific regularization to promote training effectiveness.

Our main contributions are summarized as follows: 
\begin{itemize}
    \item We identify that error accumulation from overconfident predictions in the foundation model hinders performance improvement when transferred to downstream tasks.
    \item We introduce the Self-Mutual Confidence evaluation module (SMC), which determines the integration ratio of the pseudo-labels from both models by evaluating both self-stability and mutual consistency.
    \item We design the Consensus-Divergence Consistency Regularization (CDCR) to encourage both models to make high-confidence and reliable predictions, while aligning their representation capabilities.
    \item Extensive experiments are conducted on four public multi-domain datasets, demonstrating that our method outperforms other state-of-the-art approaches\footnote{Our method also achieves competitive performance on SSMIS and UDA setting, as shown in the supplementary materials.}. With only 20 labeled data from Prostate dataset, our method obtains an improvement of over 10\% Dice than other methods.
\end{itemize}
\section{Related Work}
\label{sec:related_work}

\textbf{Medical Image Segmentation with Limited Annotation.}
Existing semi-supervised medical image segmentation (SSMIS) methods can be categorized into pseudo-label~\cite{lee2013pseudo,rizve2021defense,yang2022st++} and consistency regularization based methods~\cite{chen2021semi,cai20233d,jiang2024ph,he2024frcnet,wang2024enhancing}. Pseudo-label based methods generate pseudo-labels for unlabeled data to update the network iteratively. Wang~\etal~\cite{wang2023mcf} propose to evaluate the performance of different networks to select more reliable pseudo-labels dynamically. 
Consistency regularization based methods aim to generate consistent predictions for unlabeled data under perturbations of the input, feature, or network. Chen~\etal\cite{chen2021semi} encourage models with different initializations to produce the same predictions. 
\begin{figure*}[!t]
\centering
\includegraphics[width=0.75\linewidth]{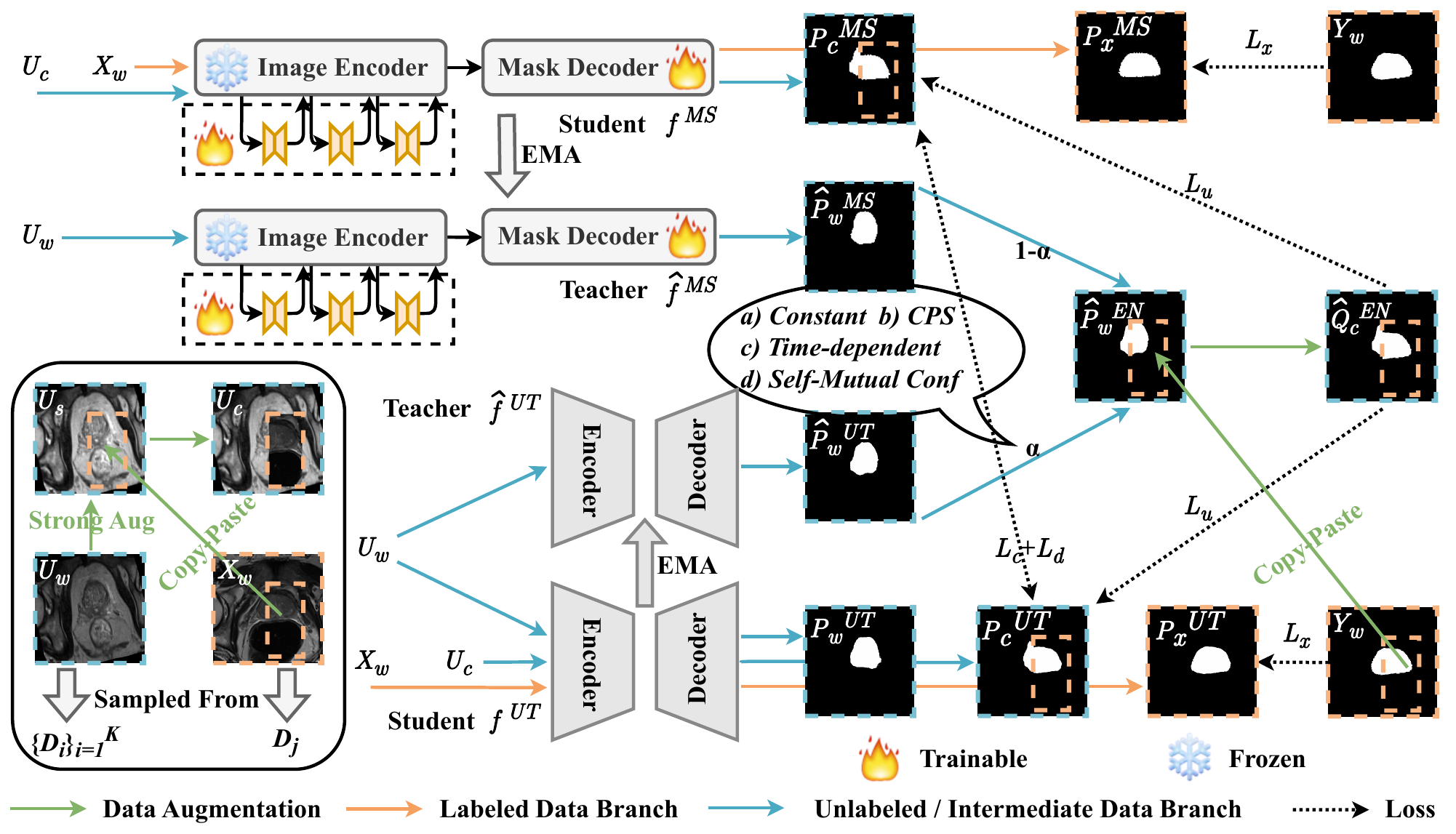}
\vspace{-5pt}
\caption{The overall framework of our SynFoC. For U-Net and MedSAM, the teacher model generates pseudo-labels for intermediate samples to guiding the student model. To reduce computational costs, we applies the LoRA module to MedSAM. We design various pseudo-label integration strategies to combine the predictions of both models, aiming to achieve higher-quality pseudo-labels. Additionally, we introduce consensus-divergence consistency regularization to enhance the efficiency of the synergistic training.}
\label{fig:framework}
\vspace{-18pt}
\end{figure*}
However, SSMIS methods typically follow the assumption that labeled and unlabeled data are sampled from the same distribution~\cite{wu2022exploring,you2022simcvd}, which can be impractical in real-world applications.

\begin{table}[t]
\centering
\resizebox{0.85\linewidth}{!}{
\begin{tabular}{c|ccc}
    \toprule
    Setting & Limited Annotations & Domain Shift & Unknown Domain Labels\\
    \midrule
    SSMIS & \checkmark & \ding{55} & -\\
    UDA & \ding{55} & \checkmark & \ding{55}\\
    LE-UDA & \checkmark & \checkmark & \ding{55}\\
    MiDSS & \checkmark & \checkmark & \checkmark\\
    \bottomrule
\end{tabular}}
\vspace{-5pt}
\caption{Various settings and challenges in clinical scenarios.}
\label{tbl:settings}
\vspace{-18pt}
\end{table}

\textbf{Medical Image Segmentation with Domain Shift.}
Domain shift is frequently encountered in real-world clinical scenarios due to differences in equipment parameters, patient populations, and disease severity. Many related problem settings, such as unsupervised domain adaptation (UDA)~\cite{chen2020unsupervised, xie2022unsupervised, shin2023sdc}, label-efficient UDA (LE-UDA)~\cite{luo2017label,zhao2022uda}, and MiDSS (Mixed Domain SSMIS)~\cite{ma2024constructing}, have been proposed. As shown in~\cref{tbl:settings}, MiDSS represents a more general scenario that confronts challenges from limited annotations,  domain shifts, and unknown domain labels, where labeled data are limited and come from a single domain, while unlabeled data are mixed from multiple domains. Other scenarios, such as SSMIS, UDA, and LE-UDA~\cite{luo2017label,zhao2022uda}, can be regarded as specific cases of this broader challenge.
Ma~\etal~\cite{ma2024constructing} propose to generate symmetric intermediate samples to fully utilize the intermediate domain information. However, conventional models often struggle to effectively transfer domain knowledge and are prone to overfitting when there is a significant domain gap between labeled and unlabeled data. To address this, we leverage the powerful feature extraction and generalization capabilities of foundation models to accelerate the early-stage training of conventional models while ensuring training quality.

\textbf{Foundation Models in Medical Image Segmentation.}
Pre-trained on massive datasets, foundation models like the Segment Anything Model (SAM)~\cite{kirillov2023segment} demonstrates exceptional generalization capability across various downstream tasks with prompts such as points and bounding boxes. 
Several efforts have been made to adapt SAM for medical images. MedSAM~\cite{ma2024segment} is fine-tuned on 1.57 million image-mask pairs, and SAM-Med2D~\cite{cheng2023sam} on 4.6 million images and 19.7 million masks from both public and private datasets.
Nonetheless, fully fine-tuning SAM is resource-intensive, and precise medical prompts generation requires expert knowledge, making the process time-consuming. To overcome these challenges, many works focus on prompt-free, efficient fine-tuning of foundation models~\cite{wu2023medical,chen2023sam,zhang2023customized,cheng2024unleashing,hu2023efficiently,li2025stitching}. SAM-adaptor and SAMed, for instance, freeze the pre-trained image encoder and employ adapter~\cite{houlsby2019parameter} or low-rank-based~\cite{hu2021lora} strategies.

Recently, many works have explored leveraging SAM for SSMIS tasks. 
For instance, In SemiSAM~\cite{zhang2023semisam}, the segmentation model generates prompt information for SAM, which in turn produces predictions that offer additional supervisory signals to the conventional model. 
Even so, static SAM fails to achieve optimal performance on specific datasets. CPC-SAM~\cite{miao2024cross} automatically generate prompts and supervision across two decoder branches, enabling effective learning from both labeled and unlabeled data. These methods typically treat foundation models as the dominant, even discarding the conventional models. However, we believe that conventional models play a key role in further boosting the performance of foundation models. 
\vspace{-5pt}
\section{Method}

\subsection{Problem Formulation and Preliminary}
\label{sec:preliminary}
Let $\mathcal{L}=\{(X_i, Y_i)\}^N_{i=1}$, and $\mathcal{U}=\{U_i\}^M_{i=1}$ denote the labeled and unlabeled data sets,
where $N$ and $M$ are their respective sizes, with $M \ge N$. $X_i, U_i \in \mathbb{R}^{W\times H\times L}$ represents the input image, and $Y_i\in\{0,1,\ldots, C\}^{W\times H}$ denotes the ground truth, where
$C$ represents the number of semantic classes, with 0 indicating the background. The training data originates from $K$ different data centers $\mathcal{D}=\{D_i\}^K_{i=1}$, where the labeled data $\mathcal{L}$ is sampled from a single domain $D_j$ and the unlabeled data $\mathcal{U}$ from multiple domains.

We first introduce the training paradigm in the MiDSS scenario. For any network structure, we define a teacher model $\hat{f}$ and a student model $f$ in the Mean Teacher architecture. Weak and strong augmentations are applied to unlabeled data $U$ to generate $U_w$ and $U_s$, respectively. 
The teacher model predicts $\hat{P}_w$ on $U_w$ and obtains the pseudo-label $\hat{Q}_w = \arg\max(\hat{P}_w)$. To bridge the domain gap, we generate an intermediate sample $U_c$ by pasting part of the weakly-augmented labeled data $X_w$ onto $U_s$. The pseudo-label $\hat{Q}_c$ for $U_c$ is obtained from $Y_w$ and $\hat{Q}_w$ similarly:
\vspace{-5pt}
\begin{equation}
    \label{eq:Copy-Paste}
    \begin{gathered}
        U_c=X_w \odot M + U_s \odot (\mathbf{1}-M),\\
        \hat{Q}_c=Y_w \odot M + \hat{Q}_w \odot (\mathbf{1}-M),
    \end{gathered}
\end{equation}
where $M \in \{0,1\}^{W \times H}$ is a one-centered mask that indicates the region for Copy-Paste. $\mathbf{1}$ represents an all-ones matrix, and $\odot$ denotes the element-wise multiplication, respectively. $\hat{Q}_c$ will be used as the supervision to supervise the student model prediction $P_c$ of $U_c$. $\hat{f}$ is updated by the Exponential Moving Average (EMA) of $f$. 

\subsection{Overview}
The overall framework of our SynFoC is illustrated in \cref{fig:framework}. Based on the training paradigm mentioned above, we incorporate the foundation model MedSAM with the lightweight model U-Net for synergistic training. The supervised training of both models is independent, with the supervised loss $L_x$ calculated based on the difference between the prediction $P_x$ of $X_w$ and $Y_w$. For unsupervised loss $L_u$, we determine the pseudo-label ensemble ratio by self and mutual confidence, providing higher-quality supervision for $U_c$. Finally, we apply specific consistency regularizations on the consistent and divergent regions of the predictions from the two student models, calculating $L_c$ and $L_d$ accordingly. The overall loss $L_{total}$ is defined as:
\begin{equation}
    L_{total} = L_x+\lambda(L_u+L_c+L_d),
\end{equation}
where $\lambda(t) = e^{-5(1-t/t_{max})}$ is a time-dependent Gaussian warming-up function. $t$ represents the current training step, and $t_{max}$ is the maximum number of steps.

\begin{figure}[t]
\centering
\includegraphics[width=0.8\linewidth]{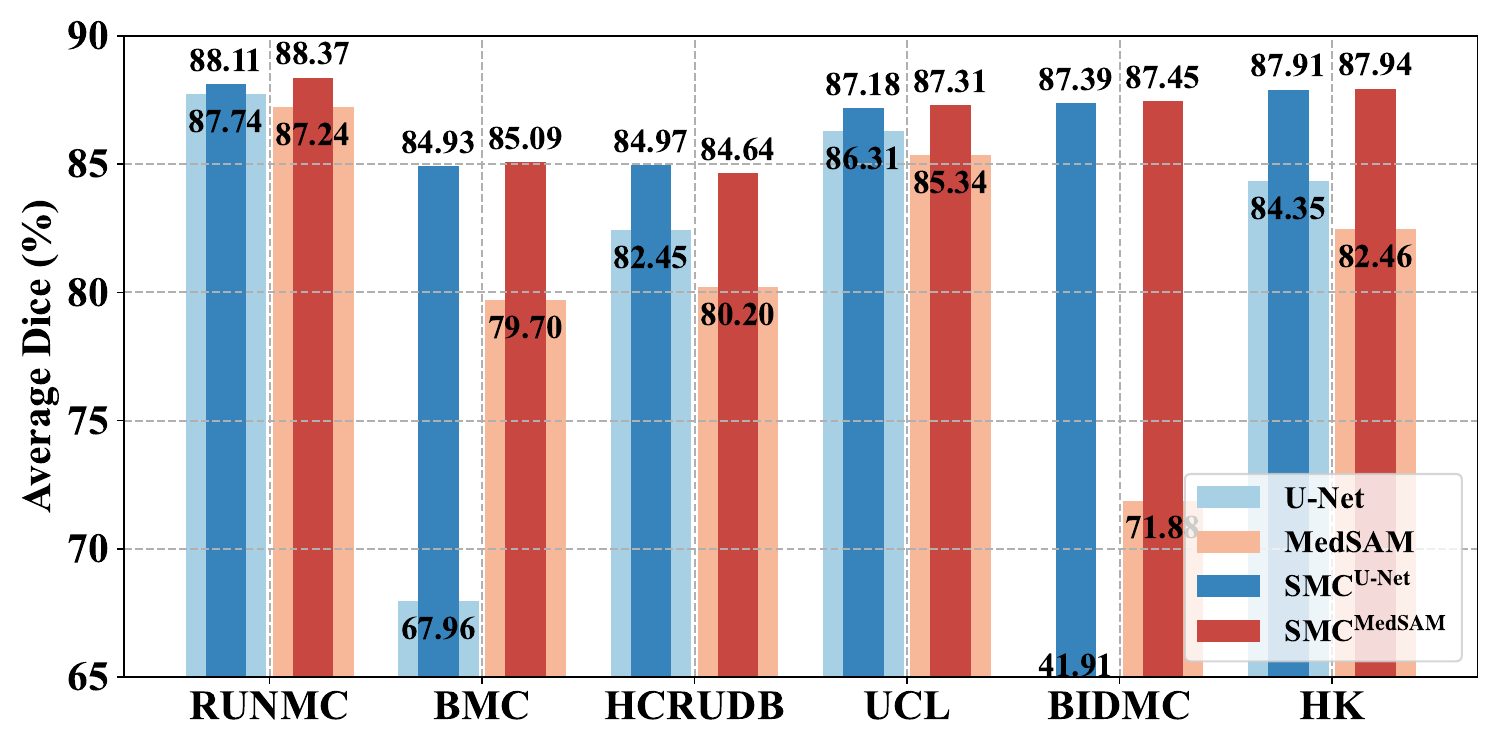}
\vspace{-5pt}
\caption{The performance comparison of U-Net and MedSAM under standalone and SMC-based synergistic training on Prostate.}
\label{fig:UNet-MedSAM-performance}
\vspace{-10pt}
\end{figure}

\begin{figure}[t]
\centering
\includegraphics[width=0.9\linewidth]{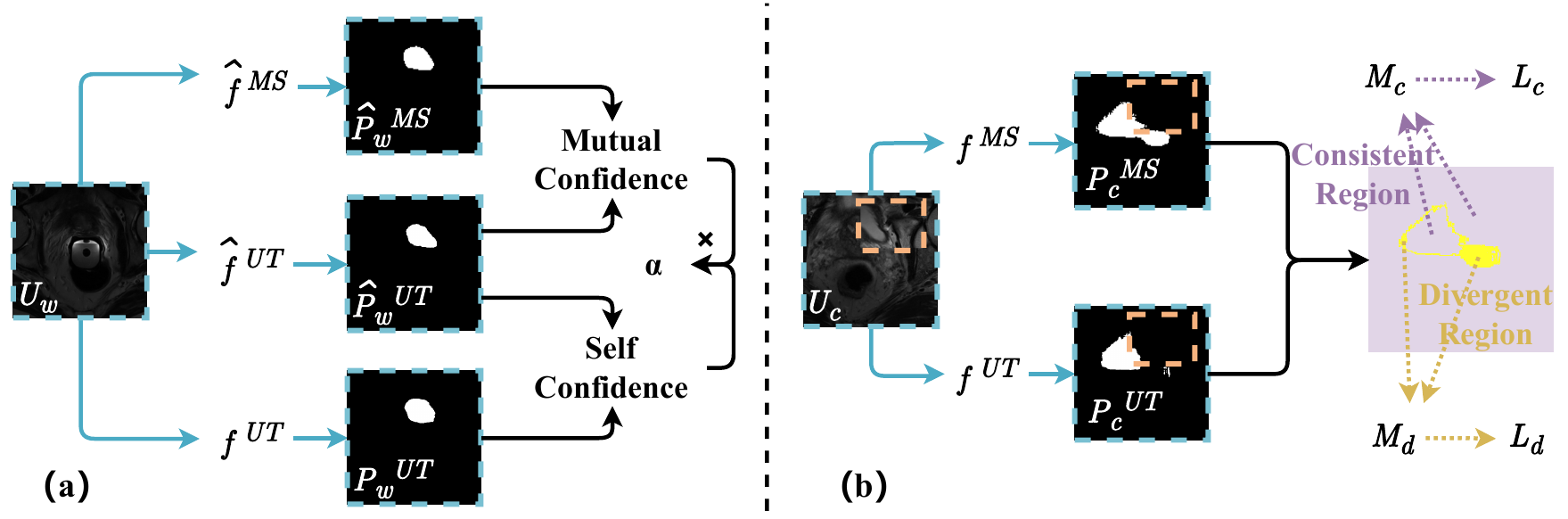}
\vspace{-5pt}
\caption{Illustration of self-confidence and mutual confidence evaluation and Consensus-Divergence consistency regularization.}
\label{fig:modules}
\vspace{-15pt}
\end{figure}


\subsection{Synergistic Training of Foundation and Conventional Models}
\label{sec:Synergistic-Training}
In the MiDSS scenario, the insufficient discriminative segmentation information from labeled data and the inefficiency of domain knowledge transfer lead to U-Net overfitting on the labeled data, as illustrated by the performance of BIDMC in~\cref{fig:UNet-MedSAM-performance}. MedSAM, with its powerful feature extraction and generalization capabilities, effectively address these shortcomings. To reduce computational cost, following SAMed~\cite{zhang2023customized} and apply Low-rank (LoRA) to the frozen image encoder, training it together with the mask decoder. The EMA update also involves only these two modules.
However, MedSAM tends to make high-confidence predictions even in the early stages of training due to its extensive prior knowledge. As demonstrated by the performance of HCRUDB in~\cref{fig:UNet-MedSAM-performance}, the high-confidence mispredictions are difficult to rectify, hindering performance improvement.

To enhance the stability of the early training for U-Net and further improve the performance of MedSAM, we propose a synergistic training strategy, which dynamically adjusts the instance-wise ensemble ratio $\alpha$ for predictions generated by both models. We denote MedSAM as $f^{MS}$ and U-Net as $f^{UT}$, with corresponding teacher models $\hat{f}^{MS}$ and $\hat{f}^{UT}$. The ensemble pseudo-label $\hat{P}^{EN}_w$ is defined as:
\begin{equation}
    \label{eq:ensemble}
    \hat{P}^{EN}_w = \alpha\hat{P}^{UT}_w + (1-\alpha)\hat{P}^{MS}_w,
\end{equation}
where $\hat{P}^{UT}_w$ and $\hat{P}^{MS}_w$ represent the pseudo-labels generated by $f^{UT}$ and $f^{MS}$ for $U_w$, respectively. 
$\alpha$ can be determined in various ways. A straightforward way is to set it as a constant, such as $\alpha = 0.5$, or to adopt a CPS strategy where $\alpha = 1$ for MedSAM and $\alpha = 0$ for U-Net. Considering the complementary characteristics between both models during training, a time-dependent linear change can be applied, \textit{i.e.,} $\alpha = t/t_{max}$. Furthermore, we incorporate instance-wise pseudo-label uncertainty and determine the ratio $\alpha$ by taking into account both self-confidence and mutual confidence (SMC), as observed in~\cref{fig:modules}(a). According to~\cref{fig:UNet-MedSAM-performance}, both models demonstrate significant performance improvement under SMC-based synergistic training.


\textbf{Self-Confidence.} It is difficult for teacher and student models to achieve consensus for hard unlabeled data~\cite{grill2020bootstrap,Wang_Xu_Liu_Zhang_Fu_2020}. High-quality pseudo-labels require models to demonstrate consistent and stable predictions. We evaluate the Self-confidence $\Phi^{self}$ of the U-Net by measuring the consistency between $P^{UT}_w$ and $\hat{P}^{UT}_w$ from the $f^{UT}$ and $\hat{f}^{UT}$:
\begin{equation}
    \Phi^{self} = \frac{1}{C}\sum_{i=1}^C\frac{2 \times (|\mathds{1}(Q^{UT}_w=i) \cap \mathds{1}(\hat{Q}^{UT}_w=i)|)}{|\mathds{1}(Q^{UT}_w=i)|+|\mathds{1}(\hat{Q}^{UT}_w=i)|},
\end{equation}
where $\mathds{1}(\cdot)$ represents the indicator function, $\mathds{1}(Q^{UT}_w=i)$ is the binary mask for pixels predicted as class $i$, and $|A|$ denotes the number of pixels where value is 1. The intersection $|\mathds{1}(Q^{UT}_w=i) \cap \mathds{1}(\hat{Q}^{UT}_w=i)|$ counts the number of pixels predicted as class $i$ by both models.

\textbf{Mutual Confidence.} When optimization converges, regardless of the performance, the model always exhibits high stability, making it difficult to ensure the quality of the pseudo-labels. Given robust feature extraction capabilities of MedSAM, it can quickly pinpoint segmentation target areas and generate reasonably accurate results. To further assess the reliability of the predictions generated by U-Net, we measure the consistency between $\hat{P}^{UT}_w$ and $\hat{P}^{MS}_w$ produced by U-Net and MedSAM to determine the mutual confidence $\Phi^{mut}$, which is defined as follows:
\begin{equation}
    \Phi^{mut} = \frac{1}{C}\sum_{i=1}^C\frac{2 \times (|\mathds{1}(\hat{Q}^{MS}_w=i) \cap \mathds{1}(\hat{Q}^{UT}_w=i)|)}{|\mathds{1}(\hat{Q}^{MS}_w=i)|+|\mathds{1}(\hat{Q}^{UT}_w=i)|}.
\end{equation}

We assess the reliability of the predictions generated by U-Net from above two perspectives. We consider 
the prediction of U-Net is reliable when both $\Phi^{self}$ and $\Phi^{mut}$ approach 1, leading to large ensemble ratio $\alpha$, and vice versa:
\begin{equation}
    \alpha=\Phi^{self}\times\Phi^{mut}.
\end{equation}

According to~\cref{eq:ensemble}, we obtain the ensembled probability map $\hat{P}_w^{EN}$ and the pseudo-label $\hat{Q}_w^{EN}$ of $U_w$. Next, we generate the intermediate sample $U_c$ along with its pseudo-label $\hat{Q}_c^{EN}$ by~\cref{eq:Copy-Paste}. $\hat{Q}_c^{EN}$ guides the predictions $P_c^{UT}$ and $P_c^{MS}$ on $U_c$, with the unsupervised loss $L_u$ defined as:
\begin{equation}
    \label{form:unsup_loss}
    \begin{gathered}
        L_{u}=L_{ce}(\hat{Q}_c^{EN}, P_c^{UT},W_c^{EN})+L_{dice}(\hat{Q}_c^{EN}, P_c^{UT},W_c^{EN})+\\
        L_{ce}(\hat{Q}_c^{EN}, P_c^{MS},W_c^{EN})+L_{dice}(\hat{Q}_c^{EN}, P_c^{MS},W_c^{EN}),
    \end{gathered}
\end{equation}
where $W_c^{EN} = \mathds{1}(\text{max}(\hat{P}_w^{EN})\ge \tau)$ represents the high-confidence regions in $\hat{Q}_c^{EN}$, and $\tau=0.95$ is a predefined confidence threshold used to filter out noisy labels. $L_{ce}$ and $L_{dice}$ denote the cross-entropy loss and dice loss, which are formulated in the supplementary materials.

\subsection{Region-Specific Consistency Regularization}
We hypothesize that the regions where both models make consistent predictions are more likely to be accurate, while the divergent regions reflect the differences in their representational capabilities. To enhance the synergistic training efficiency of both models while aligning their representational capabilities, we propose the consensus-divergence consistency regularization (CDCR). Referring to~\cref{fig:modules}(b), by comparing the predictions of the two models, we obtain consistent and divergent regions, denoted as $M_{c}$ and $M_{d}$, respectively, and apply different constraints to each region:
\begin{equation}
    \label{eq:consensus-divergence-mask}
    \begin{gathered}
        M_{c} = \mathds{1}(Q_s^{UT} = Q_s^{MS}),
        M_{d} = \mathbf{1} - M_{c}.
    \end{gathered}
\end{equation}

\textbf{Consensus Consistency Regularization} We encourage the models to generate high-confidence predictions, characterized by low-entropy probability distributions, in the areas where predictions are reliable. Specifically, we minimize the Shannon entropy for predictions in the $M_c$:
\begin{equation}
    L_{c} = -\frac{1}{S}\sum(P_c^{UT}\log P_c^{UT}+P_c^{MS}\log P_c^{MS}) \odot M_{c},
\end{equation}
where $S = W \times H \times C$.

\textbf{Divergence Consistency Regularization} We aim to reduce the prediction discrepancies to promote consistent improvement in representational capabilities. Therefore, we minimize the mean squared error (MSE) within the $M_{d}$:
\begin{equation}
    L_{d} = -\frac{1}{S}\sum\text{MSE}(P_c^{UT}, P_c^{MS}) \odot M_{d}.
\end{equation}


\begin{table}[t]
\centering
\resizebox{0.8\linewidth}{!}{
\begin{tabular}{c|ccc}
    \toprule
     & U-Net alone & MedSAM alone & SynFoC\\
    \midrule
    Training (h) & 4.09 & 7.54 & 8.61\\
    Testing (s) & 11.66 & 21.04 & 21.04\\
    \bottomrule
\end{tabular}}
\vspace{-5pt}
\caption{The training (h) and testing (s) time on Prostate dataset.}
\label{tbl:timecost}
\vspace{-18pt}
\end{table}

\label{sec:experiments}
\begin{table*}[t]
\centering
\footnotesize
\resizebox{0.9\linewidth}{!}{
\begin{tabular}{l|c|c|cccccc|cccc}
    \toprule
    \multirow{2}{*}{Methods} & \multirow{2}{*}{Venue} & \multirow{2}{*}{\#L} & \multicolumn{6}{c|}{(Prostate Segmentation) DSC $\uparrow$} & DSC $\uparrow$ & Jaccard $\uparrow$ & 95HD $\downarrow$ & ASD $\downarrow$\\
    & & & RUNMC & BMC & HCRUDB & UCL & BIDMC & HK & \multicolumn{4}{c}{Avg.}\\
    \midrule
    SupOnly & - & 20 & 22.11 & 21.81 & 19.60 & 13.87 & 18.16 & 26.98 & 20.42 & 15.63 & 118.15 & 79.70\\
    UA-MT \cite{yu2019uncertainty} & MICCAI'19 & 20 & 19.09 & 13.66 & 16.07 & 37.30 & 15.23 & 11.22 & 18.76 & 13.44 & 127.59 & 85.76\\
    FixMatch \cite{sohn2020fixmatch} & NeurIPS'20 & 20 & 81.69 & 65.27 & 53.70 & 70.40 & 10.20 & 81.22 & 60.41 & 52.53 & 49.47 & 28.83\\
    SS-Net \cite{wu2022exploring} & MICCAI'22 & 20 & 14.92 & 11.64 & 14.49 & 34.31 & 15.45 & 12.52 & 17.22 & 12.65 & 119.73 & 81.38\\
    BCP \cite{bai2023bidirectional} & CVPR'23 & 20 & 64.79 & 62.46 & 50.49 & 55.08 & 63.31 & 57.64 & 58.96 & 48.74 & 56.81 & 27.77\\
    CauSSL \cite{miao2023caussl} & ICCV23 & 20 & 20.36 & 31.11 & 15.68 & 27.27 & 26.17 & 26.66 & 24.54 & 18.03 & 116.15 & 70.57\\
    ABD \cite{chi2024adaptive} & CVPR'24 & 20 & 53.10 & 62.28 & 9.17 & 59.22 & 51.92 & 22.19 & 42.98 & 32.35 & 75.73 & 47.17\\
    SymGD \cite{ma2024constructing} & CVPR'24 & 20 & \underline{88.34} & \underline{83.26} & \underline{83.99} & \underline{85.45} & 42.03 & \underline{78.02} & \underline{76.85} & \underline{67.88} & 39.02 & 21.08\\
    \midrule
    SAMed \cite{zhang2023customized} & arXiv'23 & 20 & 63.67 & 65.62 & 65.25 & 68.65 & 48.83 & 75.31 & 64.56 & 53.49 & 35.85 & 15.83\\
    SemiSAM$^\dagger$ \cite{zhang2023semisam} & arXiv'23 & 20 & 78.07 & 67.16 & 76.49 & 69.25 & 65.97 & 69.76 & 71.12 & 59.69 & \underline{28.19} & \underline{12.61}\\
    H-SAM \cite{cheng2024unleashing} & CVPR'24 & 20 & 56.29 & 51.23 & 41.59 & 60.50 & 51.94 & 51.86 & 52.24 & 42.37 & 69.31 & 36.15\\
    CPC-SAM \cite{miao2024cross} & MICCAI'24 & 20 & 77.23 & 66.46 & 66.98 & 76.79 & \underline{79.71} & 75.42 & 73.77 & 62.85 & 31.08 & 13.16\\
    \midrule
    \color{gray} $\text{SynFoC}^\text{U-Net}$ & \color{gray} This paper & \color{gray} 20 & \color{gray} 88.09 & \color{gray} 85.22 & \color{gray} 84.97 & \color{gray} 87.74 & \color{gray} 87.63 & \color{gray} 87.74 & \color{gray} 86.90 & \color{gray} 79.11 & \color{gray} 11.25 & \color{gray} 4.82\\
    \rowcolor{Gray} $\text{SynFoC}^\text{MedSAM}$ & This paper & 20 & \textbf{88.54} & \textbf{85.74} & \textbf{84.89} & \textbf{87.51} & \textbf{87.92} & \textbf{88.34} & \textbf{87.16} & \textbf{79.30} & \textbf{10.26} & \textbf{4.41}\\
    \bottomrule
\end{tabular}}
\caption{Comparison of different methods on Prostate dataset. \#L represents the number of labeled data. 
The best performance is marked as \textbf{bold}, and the second-best is \underline{underlined}. $^\dagger$ denotes that we reproduce the results of SemiSAM.}
\label{prostate}
\vspace{-5pt}
\end{table*}
\begin{table*}[t]
\centering
\footnotesize
\resizebox{0.9\linewidth}{!}{
\begin{tabular}{l|c|c|cccc|cccc}
    \toprule
    \multirow{2}{*}{Methods} & \multirow{2}{*}{Venue} & \multirow{2}{*}{\#L} & \multicolumn{4}{c|}{(Optic Cup / Optic Disc Segmentation) DSC $\uparrow$} & DSC $\uparrow$ & Jaccard $\uparrow$ & 95HD $\downarrow$ & ASD $\downarrow$\\
    & & & Domain 1 & Domain 2 & Domain 3 & Domain 4 & \multicolumn{4}{c}{Avg.}\\
    \midrule
    SupOnly & - & 20 & 59.54 / 73.89 & 71.28 / 74.23 & 50.87 / 64.29 & 35.61 / 63.30 & 61.63 & 52.65 & 48.28 & 28.86\\
    UA-MT \cite{yu2019uncertainty} & MICCAI'19 & 20 & 59.35 / 78.46 & 63.08 / 74.45 & 35.24 / 47.73 & 36.18 / 55.43 & 56.24 & 47.00 & 48.64 & 31.35\\
    FixMatch \cite{sohn2020fixmatch} & NeurIPS'20 & 20 & 81.18 / 91.29 & 72.04 / 87.60 & 80.41 / \underline{92.95} & 74.58 / 87.07 & 83.39 & 73.48 & 11.77 & 5.60\\
    SS-Net \cite{wu2022exploring} & MICCAI'22 & 20 & 59.42 / 78.15 & 67.32 / 85.05 & 45.69 / 69.91 & 38.76 / 61.13 & 63.18 & 53.49 & 44.90 & 25.73\\
    BCP \cite{bai2023bidirectional} & CVPR'23 & 20 & 71.65 / 91.10 & 77.19 / \underline{92.00} & 72.63 / 90.77 & 77.67 / 91.42 & 83.05 & 73.66 & 11.05 & 5.80\\
    CauSSL \cite{miao2023caussl} & ICCV'23 & 20 & 63.38 / 80.60 & 67.52 / 80.72 & 49.53 / 63.88 & 39.43 / 49.43 & 61.81 & 51.80 & 41.25 & 23.94\\
    ABD \cite{chi2024adaptive} & CVPR'24 & 20 & 73.92 / 79.71 & 65.19 / 90.96 & 77.61 / 86.11 & 74.79 / 86.72 & 79.38 & 69.28 & 13.99 & 8.14\\
    SymGD \cite{ma2024constructing} & CVPR'24 & 20 & \underline{83.71} / 92.96 & 80.47 / 89.93 & \underline{84.18} / 92.97 & \underline{83.71} / \underline{93.38} & \underline{87.66} & \underline{79.10} & \underline{8.21} & \underline{3.89}\\
    \midrule
    SAMed \cite{zhang2023customized} & arXiv'23 & 20 & 71.00 / \underline{93.53} & \underline{81.77} / 90.04 & 82.07 / 92.25 & 71.62 / 93.14 & 84.47 & 75.69 & 9.83 & 5.25\\
    SemiSAM$^\dagger$ \cite{zhang2023semisam} & arXiv'23 & 20 & 83.70 / 93.21 & 72.40 / 87.72 & 81.39 / 92.11 & 79.17 / 91.10 & 85.10 & 75.50 & 9.48 & 4.60\\
    H-SAM \cite{cheng2024unleashing} & CVPR'24 & 20 & 76.97 / 93.01 & 79.01 / 90.47 & 76.85 / 91.86 & 81.03 / 92.42 & 85.20 & 76.26 & 9.67 & 4.99\\
    CPC-SAM \cite{miao2024cross} & MICCAI'24 & 20 & 75.99 / \textbf{94.34} & 80.10 / \textbf{93.08} & 83.19 / 92.81 & 83.43 / 93.20 & 87.02 & 78.65 & 8.50 & 4.24\\
    \midrule
    \color{gray} $\text{SynFoC}^\text{U-Net}$ & \color{gray} This paper & \color{gray} 20 & \color{gray} 84.26 / 92.78 & \color{gray} 78.51 / 90.12 & \color{gray} 85.33 / 93.05 & \color{gray} 82.05 / 93.88 & \color{gray} 87.50 & \color{gray} 78.93 & \color{gray} 7.74 & \color{gray} 3.79\\
    \rowcolor{Gray} $\text{SynFoC}^\text{MedSAM}$ & This paper & 20 & \textbf{85.44} / 93.29 & \textbf{82.50} / 90.52 & \textbf{85.51} / \textbf{93.56} & \textbf{83.78} / \textbf{94.23} & \textbf{88.60} & \textbf{80.50} & \textbf{6.56} & \textbf{3.47}\\
    \bottomrule
\end{tabular}}
\caption{Comparison of different methods on Fundus dataset.}
\label{fundus}
\vspace{-10pt}
\end{table*}
\begin{table*}[t]
\setlength\tabcolsep{1.0mm}
\centering
\footnotesize
\resizebox{0.9\linewidth}{!}{
\begin{tabular}{l|c|c|cccc|cccc}
    \toprule
    \multirow{2}{*}{Method} & \multirow{2}{*}{Venue} & \multirow{2}{*}{\#L} & \multicolumn{4}{c|}{(LV / MYO / RV Segmentation) DSC $\uparrow$} & DSC $\uparrow$ & Jaccard $\uparrow$ & 95HD $\downarrow$ & ASD $\downarrow$\\
    & & & Vendor A & Vendor B & Vendor C & Vendor D & \multicolumn{4}{c}{Avg.}\\
    \midrule
    SupOnly & - & 5 & 33.65 / 19.07 / 27.38 & 47.98 / 47.79 / 29.89 & 23.55 / 12.89 / 20.52 & 43.82 / 34.15 / 37.29 & 31.50 & 24.04 & 65.13 & 40.41\\
    UA-MT \cite{yu2019uncertainty} & MICCAI'19 & 5 & 15.64 / 10.57 / 10.38 & 40.07 / 35.84 / 11.64 & 17.68 / 13.30 / 12.06 & 32.32 / 19.76 / 17.90 & 19.76 & 14.53 & 78.23 & 54.74\\
    FixMatch \cite{sohn2020fixmatch} & NeurIPS'20 & 5 & 80.57 / 66.29 / 65.13 & 87.88 / \underline{79.77} / \underline{77.01} & 83.37 / 75.47 / 71.89 & 89.13 / 78.83 / \underline{78.34} & 77.81 & 67.47 & \underline{9.09} & \textbf{4.85}\\
    SS-Net \cite{wu2022exploring} & MICCAI'22 & 5 & 9.90 / 6.89 / 4.77 & 32.68 / 32.30 / 15.26 & 7.15 / 6.13 / 4.39 & 23.20 / 16.24 / 5.28 & 13.68 & 10.06 & 84.29 & 64.06\\
    BCP \cite{bai2023bidirectional} & CVPR'23 & 5 & 49.99 / 18.12 / 19.55 & 84.41 / 69.04 / 68.75 & 57.25 / 40.28 / 42.80 & 69.10 / 56.43 / 58.83 & 52.88 & 43.97 & 37.10 & 22.67\\
    CauSSL \cite{miao2023caussl} & ICCV'23 & 5 & 33.83 / 18.92 / 17.43 & 35.00 / 32.70 / 21.42 & 12.38 / 16.48 / 16.13 & 28.35 / 28.21 / 22.89 & 23.65 & 17.12 & 69.80 & 41.75\\
    ABD \cite{chi2024adaptive} & CVPR'24 & 5 & 38.74 / 24.05 / 1.56 & 29.47 / 24.20 / 17.36 & 14.62 / 7.87 / 10.85 & 39.69 / 31.30 / 35.08 & 22.90 & 17.28 & 63.48 & 51.38\\
    SymGD \cite{ma2024constructing} & CVPR'24 & 5 & 62.35 / 70.53 / 68.15 & \textbf{89.24} / \textbf{81.35} / \textbf{81.37} & 82.15 / \underline{78.32} / \underline{77.40} & \underline{89.23} / \underline{79.85} / \underline{78.00} & \underline{78.16} & \underline{67.98} & 12.91 & 6.50\\
    \midrule
    SAMed \cite{zhang2023customized} & ArXiv'23 & 5 & 66.99 / 51.21 / 28.20 & 77.67 / 60.16 / 48.80 & 72.98 / 49.72 / 37.68 & 77.72 / 55.85 / 42.36 & 55.78 & 43.61 & 35.27 & 15.38\\
    SemiSAM$^\dagger$ \cite{zhang2023semisam} & ArXiv'23 & 5 & 28.88 / 21.73 / 21.52 & 87.16 / 78.00 / 73.00 & 82.46 / 72.49 / 67.07 & 83.28 / 72.94 / 67.30 & 62.99 & 53.79 & 21.41 & 10.99\\
    H-SAM \cite{cheng2024unleashing} & CVPR'24 & 5 & 50.85 / 31.56 / 32.94 & 58.01 / 38.82 / 39.82 & 59.17 / 39.18 / 45.95 & 69.64 / 48.47 / 48.90 & 46.94 & 36.07 & 36.44 & 20.40\\
    CPC-SAM \cite{miao2024cross} & MICCAI'24 & 5 & \textbf{87.05} / \underline{74.31} / \underline{72.00} & 83.65 / 72.31 / 70.88 & \underline{85.02} / 73.73 / 74.03 & 86.28 / 74.21 / 68.19 & 76.81 & 65.70 & 12.37 & 5.13\\
    \midrule
    \color{gray} $\text{SynFoC}^\text{U-Net}$ & \color{gray} U-Net & \color{gray} 5 & \color{gray} 85.81 / 76.80 / 73.71 & \color{gray} 86.70 / 78.23 / 74.29 & \color{gray} 85.88 / 76.61 / 74.77 & \color{gray} 87.90 / 78.32 / 75.54 & \color{gray} 79.55 & \color{gray} 69.86 & \color{gray} 9.33 & \color{gray} 5.20\\
    \rowcolor{Gray} $\text{SynFoC}^\text{MedSAM}$ & MedSAM & 5 & \underline{85.65} / \textbf{76.40} / \textbf{76.18} & \underline{88.03} / 78.42 / 75.40 & \textbf{87.09} / \textbf{78.63} / \textbf{77.71} & \textbf{89.51} / \textbf{79.90} / 77.13 & \textbf{80.84} & \textbf{70.94} & \textbf{8.15} & \textbf{3.65}\\
    \bottomrule
\end{tabular}}
\caption{Comparison of different methods on M\&Ms dataset.}
\label{mnms}
\vspace{-15pt}
\end{table*}
\begin{table}[t]
\setlength\tabcolsep{1.0mm}
\centering
\footnotesize
\resizebox{0.9\linewidth}{!}{
\begin{tabular}{l|c|cc|cccc}
    \toprule
    \multirow{2}{*}{Method} & \multirow{2}{*}{Venue} & \multicolumn{2}{c|}{(Cancer) DSC $\uparrow$} & DSC $\uparrow$ & Jaccard $\uparrow$ & 95HD $\downarrow$ & ASD $\downarrow$\\
    & & Benign & Malignant & \multicolumn{4}{c}{Avg.}\\
    \midrule
     \multicolumn{8}{c}{64 (1/8) labels}\\
    \midrule
    SupOnly & - & 55.38 & 63.51 & 59.45 & 48.94 & 61.10 & 25.44\\
    UA-MT \cite{yu2019uncertainty} & MICCAI'19 & 53.51 & 62.68 & 58.10 & 47.96 & 55.50 & 26.56\\
    FixMatch \cite{sohn2020fixmatch} & NeurIPS'20 & 59.49 & 69.80 & 64.65 & 54.60 & 46.48 & 20.68\\
    SS-Net \cite{wu2022exploring} & MICCAI'22 & 56.11 & 63.36 & 59.74 & 49.50 & 51.30 & 22.66\\
    BCP \cite{bai2023bidirectional} & CVPR'23 & 60.49 & 65.20 & 62.85 & 52.68 & 50.80 & 18.62\\
    CauSSL \cite{miao2023caussl} & ICCV'23 & 49.54 & 59.31 & 54.43 & 44.26 & 57.09 & 29.05\\
    ABD \cite{chi2024adaptive} & CVPR'24 & 50.45 & 62.71 & 56.58 & 47.03 & 49.40 & 23.27\\
    SymGD \cite{ma2024constructing} & CVPR'24 & 60.04 & \underline{72.78} & 66.41 & 56.45 & \underline{40.26} & 18.20\\
    \midrule
    SAMed \cite{zhang2023customized} & Arxiv'23 & 66.89 & 63.52 & 65.21 & 54.13 & 46.47 & 18.39\\
    SemiSAM$^\dagger$ \cite{zhang2023semisam} & Arxiv'23 & 60.65 & 66.35 & 63.50 & 53.25 & 50.43 & 23.46\\
    H-SAM \cite{cheng2024unleashing} & CVPR'24 & 67.76 & 63.87 & 65.82 & 54.99 & 41.58 & 17.25\\
    CPC-SAM \cite{miao2024cross} & MICCAI'24 & \textbf{71.87} & 65.86 & \underline{68.87} & \underline{57.46} & 40.77 & \underline{16.29} \\
    \midrule
    \color{gray} $\text{SynFoC}^\text{U-Net}$ & \color{gray} This paper & \color{gray} 59.56 & \color{gray} 72.69 & \color{gray} 66.13 & \color{gray} 56.41 & \color{gray} 43.52 & \color{gray} 20.67\\
    \rowcolor{Gray} $\text{SynFoC}^\text{MedSAM}$ & This paper & \underline{70.16} & \textbf{73.05} & \textbf{71.61} & \textbf{61.31} & \textbf{34.77} & \textbf{15.05}\\
    \midrule
     \multicolumn{8}{c}{129 (1/4) labels}\\
    \midrule
    SupOnly & - & 59.57 & 66.05 & 62.81 & 52.31 & 54.26 & 23.29\\
    UA-MT \cite{yu2019uncertainty} & MICCAI'19 & 56.54 & 64.92 & 60.73 & 50.69 & 50.29 & 23.15\\
    FixMatch \cite{sohn2020fixmatch} & NeurIPS'20 & 61.28 & 71.13 & 66.21 & 55.77 & 49.73 & 21.28\\
    SS-Net \cite{wu2022exploring} & MICCAI'22 & 56.94 & 64.18 & 60.56 & 51.09 & 49.69 & 21.32\\
    BCP \cite{bai2023bidirectional} & CVPR'23 & 61.96 & 67.21 & 64.59 & 53.68 & 55.82 & 22.87\\
    CauSSL \cite{miao2023caussl} & ICCV'23 & 58.97 & 62.57 & 60.77 & 50.47 & 48.05 & 21.46\\
    ABD \cite{chi2024adaptive} & CVPR'24 & 56.62 & 63.85 & 60.24 & 49.34 & 48.98 & 20.83\\
    SymGD \cite{ma2024constructing} & CVPR'24 & 61.68 & \underline{72.09} & 66.89 & 56.79 & 44.23 & 17.82\\
    \midrule
    SAMed \cite{zhang2023customized} & Arxiv'23 & 70.79 & 68.09 & 69.44 & 58.40 & 38.49 & 14.67\\
    SemiSAM$^\dagger$ \cite{zhang2023semisam} & Arxiv'23 & 61.61 & 68.25 & 64.93 & 55.12 & 48.10 & 19.08\\
    H-SAM \cite{cheng2024unleashing} & CVPR'24 & 70.86 & 65.55 & 68.21 & 57.26 & 38.83 & 18.12\\
    CPC-SAM \cite{miao2024cross} & MICCAI'24 & \textbf{74.01} & 71.13 & \underline{72.57} & \underline{61.80} & \underline{34.04} & \underline{13.30}\\
    \midrule
    \color{gray} $\text{SynFoC}^\text{U-Net}$ & \color{gray} This paper & \color{gray} 61.28 & \color{gray} 71.45 & \color{gray} 66.37 & \color{gray} 56.45 & \color{gray} 49.89 & \color{gray} 22.58\\
    \rowcolor{Gray} $\text{SynFoC}^\text{MedSAM}$ & This paper & \underline{73.74} & \textbf{75.75} & \textbf{74.75} & \textbf{64.90} & \textbf{31.29} & \textbf{12.45}\\
    \bottomrule
\end{tabular}}
\caption{Comparison of different methods on BUSI dataset.}
\label{busi}
\vspace{-15pt}
\end{table}
\begin{table*}[t]
\centering
\footnotesize
\resizebox{0.90\linewidth}{!}{
\begin{tabular}{cccc|cccccc|cccc}
    \toprule
    \multirow{2}{*}{Base} & \multirow{2}{*}{SMC} & \multirow{2}{*}{CDCR} & \multirow{2}{*}{CR} & \multicolumn{6}{c|}{DSC $\uparrow$} & DSC $\uparrow$ & Jaccard $\uparrow$ & 95HD $\downarrow$ & ASD $\downarrow$\\
    & & & & RUNMC & BMC & HCRUDB & UCL & BIDMC & HK & Avg. & Avg. & Avg. & Avg.\\
    \midrule
    $\checkmark_\text{U-Net}$ & & & & 87.74 & 67.96 & 82.45 & 86.31 & 41.91 & 84.35 & 75.12 & 65.76 & 54.67 & 29.08\\
    $\checkmark_\text{MedSAM}$ & & & & 87.24 & 79.70 & 80.20 & 85.34 & 71.88 & 82.46 & 81.14 & 71.91 & 22.78 & 9.13\\
    \checkmark & \checkmark & & & 88.37 & 85.09 & 84.64 & 87.31 & 87.45 & 87.94 & 86.80 & 78.85 & 10.78 & 4.64\\
    \checkmark & & & \checkmark & 87.99 & 81.84 & 82.99 & 86.18 & 71.92 & 86.34 & 82.88 & 73.81 & 17.09 & 7.21\\
    \checkmark & & \checkmark & & 88.20 & 84.86 & 83.36 & 86.75 & 87.15 & 86.82 & 86.19 & 77.94 & 11.91 & 5.03\\
    \midrule
    \checkmark & \checkmark & \checkmark & & \textbf{88.54} & \textbf{85.74} & \textbf{84.89} & \textbf{87.51} & \textbf{87.92} & \textbf{88.34} & \textbf{87.16} & \textbf{79.30} & \textbf{10.26} & \textbf{4.41}\\
    \bottomrule
\end{tabular}
}
\vspace{-5pt}
\caption{Ablation experiments on Prostate dataset.}
\label{ablation_prostate}
\vspace{-10pt}
\end{table*}

\begin{table}[t]
\centering
\footnotesize
\resizebox{0.78\linewidth}{!}{
\begin{tabular}{c|cccc}
    \toprule
    Strategy & DSC $\uparrow$ & Jaccard $\uparrow$ & 95HD $\downarrow$ & ASD $\downarrow$\\
    \midrule
    $\text{Constant}_{0.5}$ & 85.93 & 77.75 & 11.78 & 5.06\\
    CPS & 84.30 & 75.36 & 12.59 & 5.75\\
    Linear & 86.19 & 78.12 & 11.21 & 4.85\\
    Self-only & 86.23 & 78.16 & 11.15 & 4.85\\
    Mutual-only & 86.29 & 78.18 & 11.09 & 4.76\\
    \midrule
    SMC & \textbf{86.80} & \textbf{78.85} & \textbf{10.78} & \textbf{4.64}\\
    \bottomrule
\end{tabular}
}
\vspace{-5pt}
\caption{Ablation study of different synergistic strategies.}
\label{ablation_prostate_synergistic}
\vspace{-20pt}
\end{table}

\vspace{-10pt}
\subsection{Remarks}
To maintain the decent image resolution of predicted segmentation prediction, the input image of MedSAM is upsampled from $W \times H \times L$ to $512 \times 512 \times L$~\cite{zhang2023customized}. The output resolution of the segmentation logits for each class is $128 \times 128$, which differs from that of U-Net $(W \times H)$. Interpolation is used to align the resolutions when matrix calculations involve different resolutions. During the testing phase, We retain only the student MedSAM model and resize the segmentation results to $W \times H$ for inference. We present the training and testing times in~\cref{tbl:timecost}. On Prostate dataset, our method takes approximately 1 extra hour compared to training MedSAM alone. We believe this additional time is worthwhile due to the significant improvements in training stability and model performance. During the testing phase, we require about 10 extra seconds compared to U-Net, which is acceptable. SynFoC is a general method that achieve competitive performance in traditional SSMIS and UDA settings, facilitating breakthroughs for foundation model, not limited to MedSAM, in downstream tasks. Further details are in the supplementary materials.
\section{Experiments}
\subsection{Datasets and Evaluation Metrics}
\textbf{ The Prostate dataset}~\cite{liu2020shape} is a well-organized multi-site dataset for prostate MRI segmentation, consisting of T2-weighted MRI data collected from six different data sources across three public datasets. For each domain, the data is split into training and validation sets in a 4:1 ratio. Each 2D slice is resized to a resolution of $384 \times 384$ pixels.

\noindent \textbf{The Fundus dataset}~\cite{wang2020dofe} consists of fundus images from four different medical centers, used for the segmentation of the optic cup and optic disc. The data from each domain has been pre-split into training and test sets, with an 800×800 region of interest (ROI) cropped from each image. We resize the images to $256 \times 256$ for processing.

\noindent \textbf{The M\&Ms dataset}~\cite{campello2021multi} is collected from four different magnetic resonance scanner vendors, with annotations available only for the end-systole and end-diastole phases. We split annotated data of each vendor into training and test sets at a 4:1 ratio for the segmentation tasks of the left ventricle (LV), left ventricle myocardium (MYO), and right ventricle (RV). Each slice is resized to $288 \times 288$.

\noindent \textbf{The BUSI dataset}~\cite{al2020dataset} includes breast ultrasound images categorized into three classes based on breast cancer: normal, benign, and malignant. Since there is no segmentation target for the normal class, we divide the dataset into two domains based on tumor type: benign and malignant. For each domain, the data is split into training and test sets at a 4:1 ratio. Each image is resized to $256 \times 256$.

We also evaluate our SynFoC in SSMIS and UDA on the \textbf{ACDC}~\cite{bernard2018deep} and \textbf{MS-CMRSeg}~\cite{zhuang2018multivariate} datasets, respectively. The evaluation metrics include the Dice Similarity Coefficient (DSC), Jaccard Index, 95\% Hausdorff Distance (95HD), and Average Surface Distance (ASD).

\subsection{Implementation Details}
Our method is implemented by Pytorch and trained on an NVIDIA GeForce RTX 3090 GPU. For U-Net, we use Stochastic Gradient Descent (SGD) with a momentum of 0.9 and a weight decay of 0.0001, with an initial learning rate of 0.03. For MedSAM, we adopt the ViT\_B version, where the input is resized to $512 \times 512$, and the LoRA rank is set to 4. MedSAM is optimized with the AdamW optimizer with an initial learning rate of 0.0001, $\beta_1 = 0.9$, $\beta_2 = 0.999$, and a weight decay of 0.1. We experiment with the following numbers of labeled data for each dataset: 20 for prostate and fundus, 5 for M\&MS, and 64 (1/8) or 129 (1/4) for BUSI. Except for M\&MS, the batch size is set to 8, with an equal split between labeled and unlabeled samples. The batch size for M\&MS is reduced to 4 due to the extremely small number of labeled data (only 5). The maximum number of training iterations is set to 60,000 for Prostate and M\&Ms datasets, and 30,000 for Fundus and BUSI datasets. For each dataset, taking Prostate dataset as an example, we use 20 labeled data from a specific domain, such as RUNMC, as the labeled data, while the remaining data serves as the unlabeled data. The performance is evaluated on multiple domain test sets, and the average performance across all six domains is reported as the experimental results, corresponding to the values in the column for RUNMC in~\cref{prostate}. We compare our approach with various methods based on the conventional model (UA-MT~\cite{yu2019uncertainty}, FixMatch~\cite{sohn2020fixmatch,upretee2022fixmatchseg}, SS-Net~\cite{wu2022exploring}, BCP~\cite{bai2023bidirectional}, CauSSL~\cite{miao2023caussl}, ABD~\cite{chi2024adaptive}, SymGD~\cite{ma2024constructing}) and those based on the foundation model (SAMed~\cite{zhang2023customized}, SemiSAM~\cite{zhang2023semisam}, H-SAM~\cite{cheng2024unleashing}, CPC-SAM~\cite{miao2024cross}). To ensure a fair comparison, we select U-Net as the conventional model and MedSAM as the foundation model. Among the methods, SupOnly indicates the performance of the model trained solely on labeled data.

\subsection{Comparison with State-of-the-Art Methods}
\begin{figure}[t]
\centering
\includegraphics[width=0.9\linewidth]{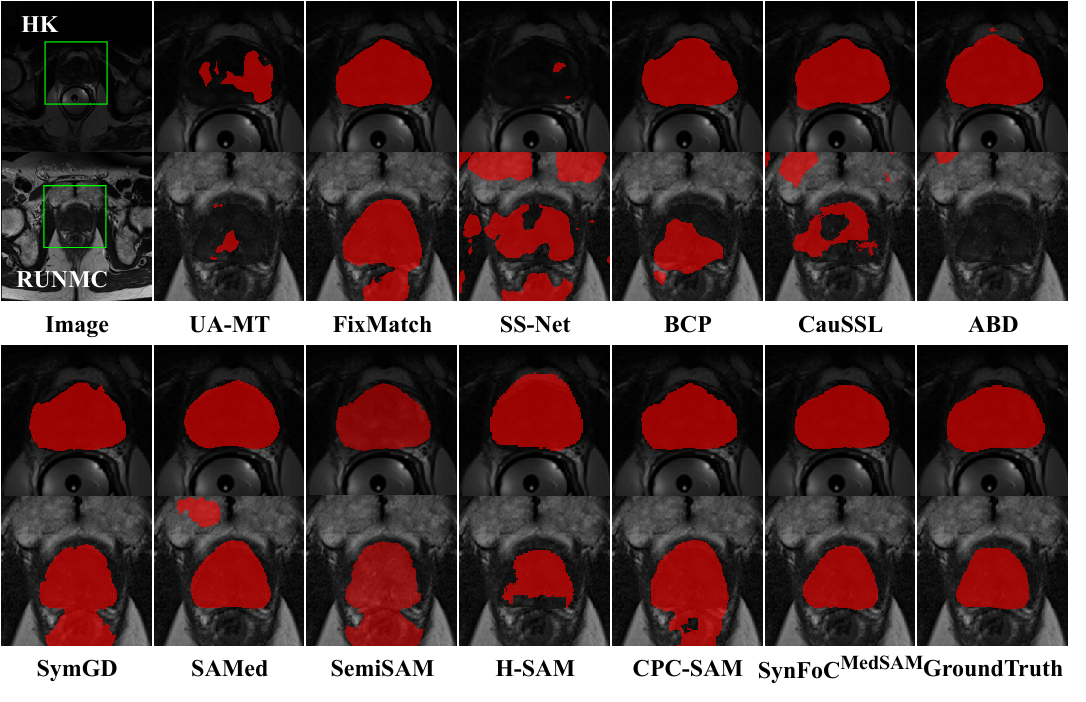}
\caption{Visual comparison of different methods on Prostate dataset. The test samples are drawn from the labeled domain (HK) and another domain (RUNMC), respectively.}
\label{fig:prostate_img}
\vspace{-15pt}
\end{figure}

\textbf{The Prostate dataset.}
\cref{prostate} presents the performance of different methods on Prostate dataset using 20 labeled data. Our method outperforms other state-of-the-art methods by a large margin, achieving an improvement of 10.31\% in DSC. For the test data sampled from the same and different domains as the labeled data, we provide visual comparisons in~\cref{fig:prostate_img} to validate the superiority of our method.

\noindent \textbf{The Fundus dataset.}
As shown in~\cref{fundus}, we conduct experiments on Fundus dataset using 20 labeled data. The segmentation performance for the optic cup and disc is separated by a slash. Through the synergistic training of U-Net and MedSAM, both models complement shortcomings of each other and enhance their performance mutually. Our method surpasses all other approaches across four metrics,  achieving state-of-the-art results. Visual comparisons on Fundus dataset are presented in the supplementary materials, and the same applies to M\&Ms and BUSI datasets.

\noindent \textbf{The M\&Ms dataset.}
In~\cref{mnms}, we evaluate the performance of various methods on M\&Ms dataset using only 5 labeled data. The segmentation results for the LV, MYO, and RV are separated by slashes. Despite the extremely limited number of labeled data, our method still achieves a 2.14\% improvement in DSC compared to other methods. 

\noindent \textbf{The BUSI dataset.}
We further examine our method on BUSI dataset in ~\cref{busi}. When 12.5\% and 25\% of labeled data are available, our SynFoC outperforms other methods by 2.74\% and 2.18\% in DSC, respectively.

\subsection{Ablation Studies}
We conduct ablation studies to verify the effectiveness of each module in our method. All experiments are conducted on Prostate dataset with 20 labeled data. 

\textbf{Effectiveness of each module.} As shown in~\cref{ablation_prostate}, \textbf{Base} refers to the SSMIS method described in~\cref{sec:preliminary}, where intermediate samples are generated to promote the training process. $\checkmark_\text{U-Net}$ and $\checkmark_\text{MedSAM}$ represent the \textbf{Base} method employing U-Net and MedSAM as the backbone models, respectively. 
Compared to the foundation model, the well-trained conventional model achieves a higher performance ceiling. However, when there is a significant discrepancy between labeled data and unlabeled data, the conventional model suffers from error accumulation, leading to training failure, whereas the foundation model continues to demonstrate strong segmentation performance. \textbf{SMC} denotes the synergistic training of U-Net and MedSAM, where the pseudo-label ensemble weight is determined based on self-mutual confidence. \textbf{CDCR} refers to the introduction of consensus-divergence consistency regularization between the two models. Both methods significantly enhance the training effectiveness of the models. In contrast to \textbf{CR}, which directly minimizes the MSE loss between $P_c^{UT}$ and $P_c^{MS}$, \textbf{CDCR} reduces the uncertainty in the consistent predictions of the two models, accelerating a more reliable convergence. Incorporating both \textbf{SMC} and \textbf{CDCR} improves performance, yielding the best results. 

\textbf{Different synergistic strategies.}
$\alpha$ can be determined through various approaches, such as \textbf{$\text{constant}_{0.5}$}, \textbf{CPS}, \textbf{linear}, and \textbf{SMC} mentioned in~\cref{sec:Synergistic-Training}. We also evaluate the performance of employing self-confidence and mutual confidence individually. As shown in~\cref{ablation_prostate_synergistic}, our \textbf{SMC} assesses the reliability of U-Net pseudo-labels at the instance level from multiple perspectives, dynamically adjusting the ensemble ratio to achieve optimal performance.

\begin{figure}[t]
\centering
\includegraphics[width=0.9\linewidth]{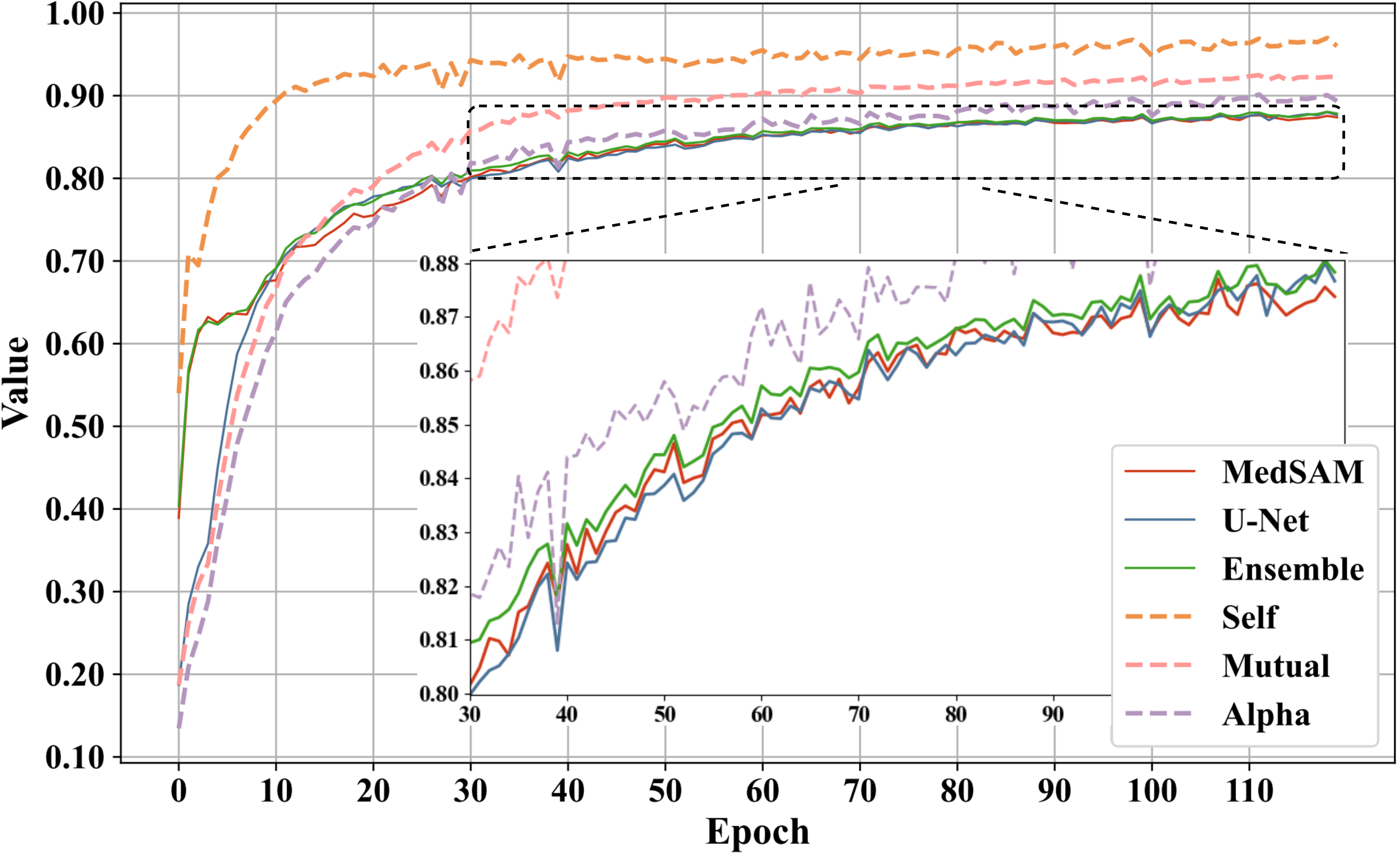}
\vspace{-5pt}
\caption{The variation of parameters alongside the quality of pseudo-labels, with 20 labeled data from the BMC domain. The dashed lines represent the curves for self-confidence, mutual confidence, and $\alpha$, respectively. The solid lines indicate the quality of pseudo-labels generated by MedSAM, U-Net, and the ensemble pseudo-labels, measured by the DSC against the ground truth.}
\label{fig:PL_para_curve}
\vspace{-15pt}
\end{figure}

\textbf{Alpha Variation and Pseudo-Label Quality.} In~\cref{fig:PL_para_curve}, self-confidence clearly reflects the stability of U-Net, as observed in the fluctuations during epochs 38 to 40, while mutual confidence roughly indicates the quality of the pseudo-labels from U-Net. The $\alpha$ more precisely captures the relationship between the quality of U-Net's pseudo-labels and its stability. The inset further illustrates that the quality of ensemble-generated pseudo-labels consistently surpasses that of the individual model outputs.
\vspace{-5pt}
\section{Conclusion}
In this paper, we propose a novel synergistic training framework for foundation and conventional models that complements shortcomings of each other. We precisely measure the quality of the pseudo-labels from U-Net by self-mutual confidence, designing the instance-wise pseudo-label ensemble ratio to generate higher-quality pseudo-labels. Furthermore, we introduce the consensus-divergence consistency regularization to ensure consistent improvement in the representation capabilities of both models and promote reliable convergence during training. Extensive experiments conducted on four public multi-domain datasets validates the effectiveness of our method.

\noindent\textbf{Acknowledgements.} 
This work was supported by NSFC Project (62222604, 62206052, 624B2063), China Postdoctoral Science Foundation (2024M750424), Fundamental Research Funds for the Central Universities (020214380120, 020214380128), State Key Laboratory Fund (ZZKT2024A14), Postdoctoral Fellowship Program of CPSF (GZC20240252), Jiangsu Funding Program for Excellent Postdoctoral Talent (2024ZB242), Jiangsu Science and Technology Major Project (BG2024031), Shandong Natural Science Foundation (ZR2023MF037), and Open Project Program of the State Key Laboratory of CAD\&CG (A2320), Zhejiang University.
\clearpage
\setcounter{page}{1}
\maketitlesupplementary

\renewcommand{\thesection}{\Alph{section}}
\setcounter{section}{0}

\section{Loss Function Formulations}
We provide the specific definitions of $L_{ce}$ and $L_{dice}$ mentioned in~\cref{form:unsup_loss} (\cref{sec:Synergistic-Training}):
\begin{equation}
    \label{ce_loss}
    L_{ce}(y,p,w)=-\frac{1}{H \times W}\sum_{i=1}^{H \times W} w_i y_i\log p_i,
\end{equation}
\begin{equation}
    \label{dice_loss}
    L_{dice}(y,p,w)=1-\frac{2\times \sum_{i=1}^{H \times W}w_i p_i y_i}{\sum_{i=1}^{H \times W} w_i(p_i^2+y_i^2)},
\end{equation}
where $y_i$, $p_i$, and $w_i$ is $i^{th}$ pixel of $y$, $p$, and $w$, respectively.

\section{Visual Results Across Multiple Datasets}
\begin{figure}[h]
\centering
\includegraphics[width=0.9\linewidth]{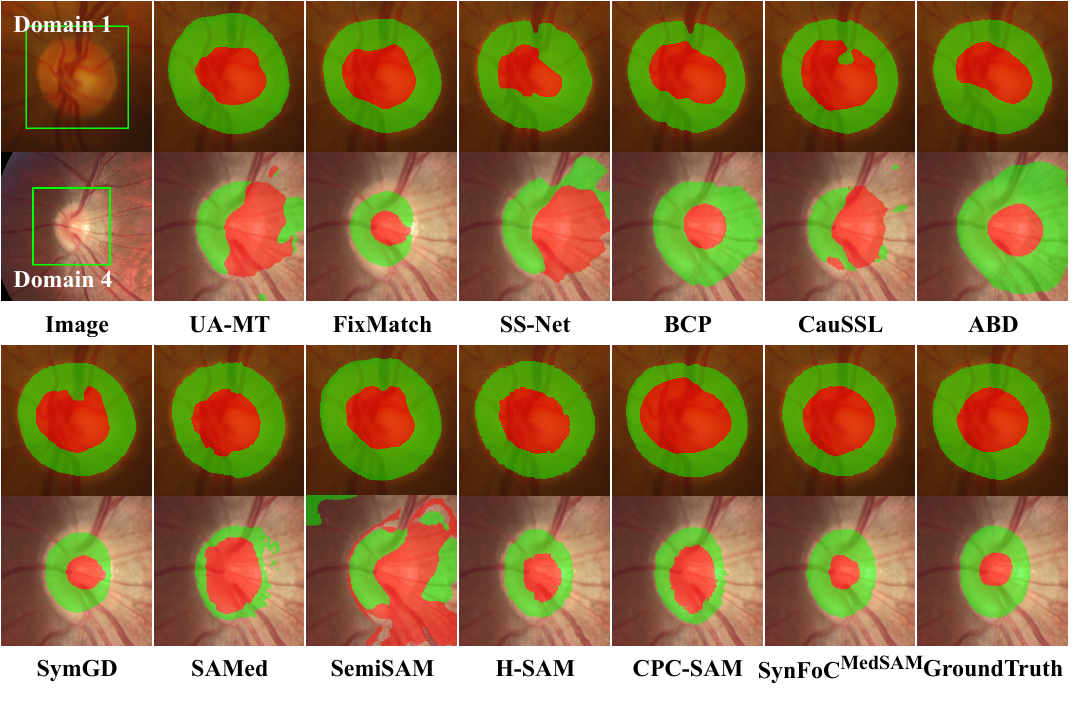}
\caption{Visual comparison of different methods on Fundus dataset. The test samples are drawn from the labeled domain 1 and another domain 4, respectively.}
\label{fig:fundus_img}
\vspace{-20pt}
\end{figure}

\begin{figure}[h]
\centering
\includegraphics[width=0.9\linewidth]{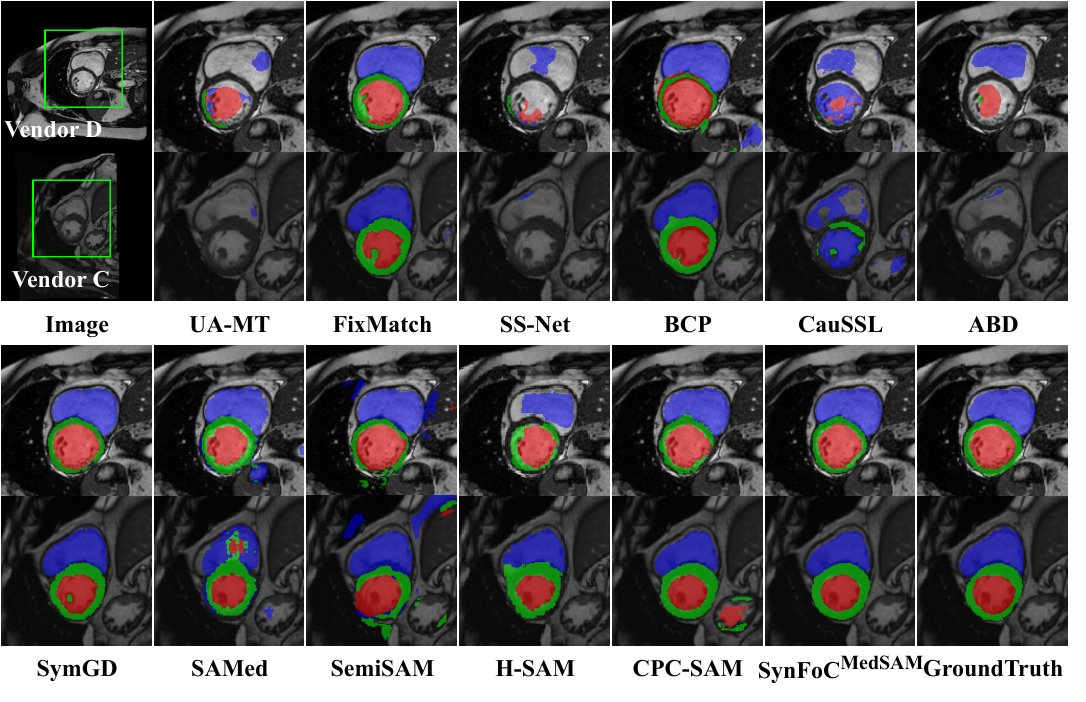}
\caption{Visual comparison of different methods on M\&Ms dataset. The test samples are drawn from the labeled Vendor D and another Vendor C, respectively.}
\label{fig:MNMS_img}
\end{figure}

\begin{figure}[h]
\centering
\includegraphics[width=0.9\linewidth]{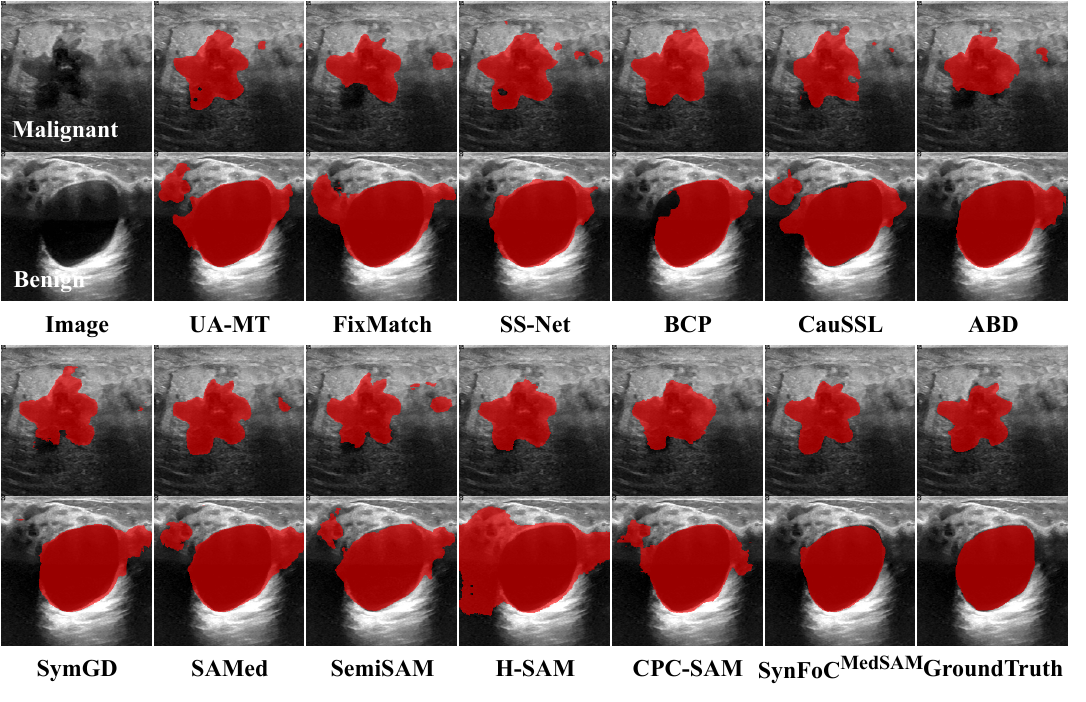}
\caption{Visual comparison of different methods on BUSI dataset. The test samples are drawn from the labeled domain Malignant and another domain Benign, respectively.}
\label{fig:BUSI_img}
\vspace{-10pt}
\end{figure}

We provide visual results of different methods on Fundus, M\&Ms, and BUSI datasets. As shown in~\cref{fig:fundus_img,fig:MNMS_img,fig:BUSI_img}, our SynFoC achieves optimal segmentation results on both test samples from the same and different domains as labeled data, with minimal error compared to the ground truth.

\section{Reproduction of SemiSAM Method}
We compare all other methods by their official code implementations, whereas we reproduce SemiSAM since its public code has not been released. SemiSAM, based on the SSMIS framework, utilize predictions from conventional model to generate prompts for frozen foundation model. In turn, foundation model generates predictions based on the prompts to provide additional supervision for conventional model. To address domain shift issue, we replace the UAMT used in the original paper with the training method described in~\cref{sec:preliminary}. We standardize the use of MedSAM as the foundation model. Since MedSAM is fine-tuned on large-scale medical data with bounding boxes based on SAM, we provide bounding box prompts from the conventional model to the frozen MedSAM in SemiSAM.

\section{Comparison with More Methods}
\begin{table}[t]
\centering
\footnotesize
\begin{tabular}{c|cccc}
    \toprule
    Method & DSC $\uparrow$ & Jaccard $\uparrow$ & 95HD $\downarrow$ & ASD $\downarrow$\\
    \midrule
     \multicolumn{5}{c}{Prostate 20 labels}\\
    \midrule
    SIFA~\cite{chen2020unsupervised} & 59.33 & 45.40 & 53.90 & 24.29\\
    UDA-VAE++~\cite{lu2022unsupervised} & 64.36 & 50.27 & 33.14 & 15.11\\
    $\text{MedSAM}^\text{Bounding box}$ & 77.39 & 63.67 & 13.22 & 6.27\\
    $\text{MedSAM}^\text{Full Fine-tuning}$ & 81.78 & 72.21 & 30.78 & 13.73\\
    SynFoC & \textbf{87.16} & \textbf{79.30} & \textbf{10.26} & \textbf{4.41}\\
    \midrule
     \multicolumn{5}{c}{Fundus 20 labels}\\
     \midrule
    SIFA~\cite{chen2020unsupervised} & 67.78 & 54.77 & 20.16 & 10.93\\
    UDA-VAE++~\cite{lu2022unsupervised} & 73.51 & 61.40 & 17.60 & 9.86\\
    $\text{MedSAM}^\text{Bounding box}$ & 77.82 & 64.87 & 15.21 & 6.62\\
    $\text{MedSAM}^\text{Full Fine-tuning}$ & 85.99 & 77.18 & 9.04 & 4.48\\
    SynFoC & \textbf{88.60} & \textbf{80.50} & \textbf{6.56} & \textbf{3.47}\\
    \bottomrule
\end{tabular}
\caption{Comparison of different methods on Prostate and Fundus datasets.}
\label{supp_comparison}
\vspace{-10pt}
\end{table}

As shown in~\cref{supp_comparison}, we conduct further comparisons on Prostate and Fundus datasets to demonstrate the superiority of our method. The methods include UDA approaches (SIFA~\cite{chen2020unsupervised} and UDA-VAE++~\cite{lu2022unsupervised}), MedSAM with precise bounding box prompts, and fully fine-tuned MedSAM (instead of LoRA-based strategy). It can be observed that UDA methods struggle to achieve satisfactory performance when the number of labeled data is limited. Despite being provided with precise bounding box prompts, MedSAM still falls short in specific datasets. Even with full fine-tuning, MedSAM struggles to correct high-confidence mispredictions, while significantly increasing training costs.

\section{More Ablation Studies}
\begin{figure}[h]
\centering
\includegraphics[width=\linewidth]{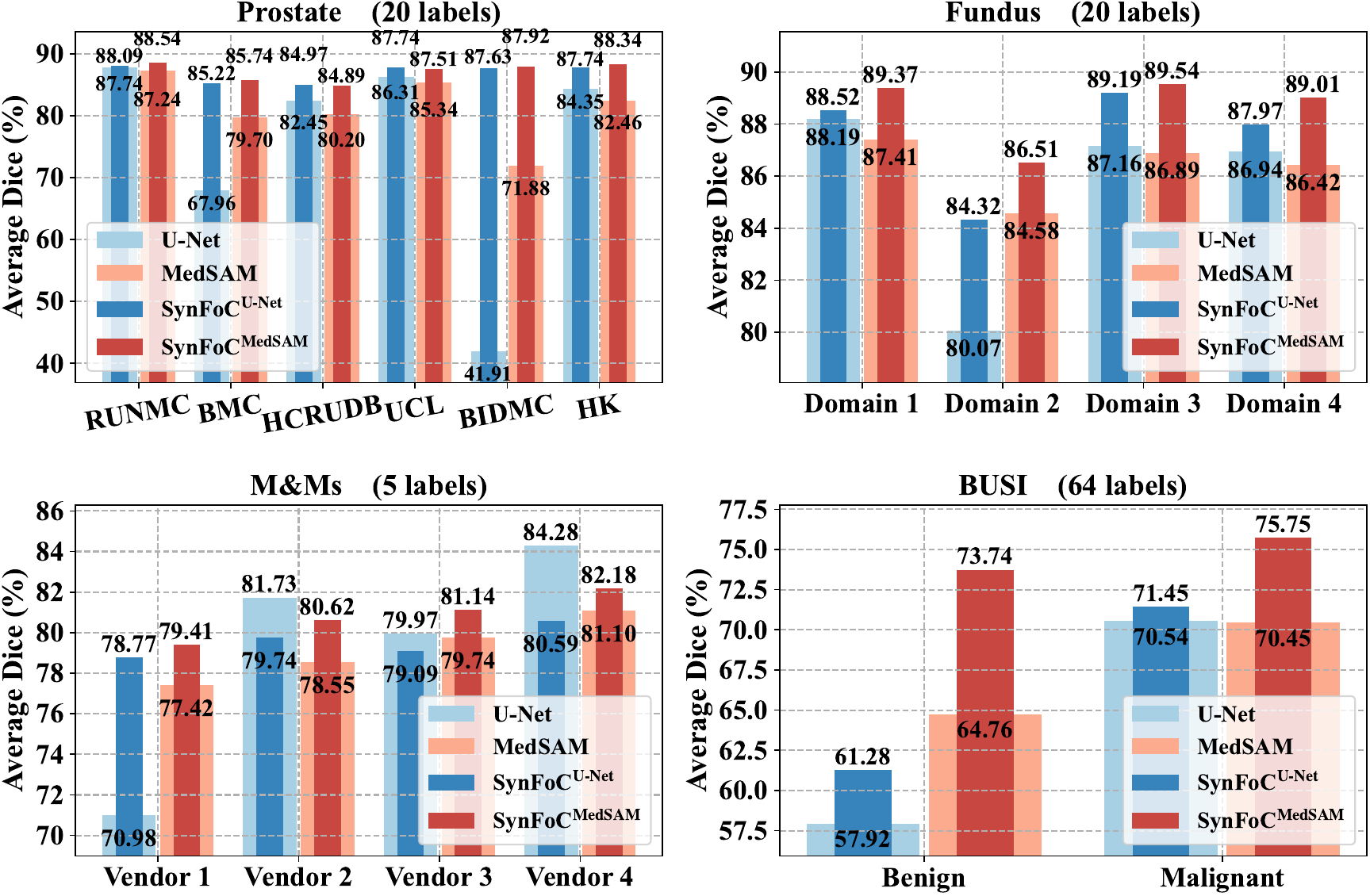}
\caption{The performance comparison of U-Net and MedSAM under standalone training and our SynFoC across four datasets.}
\label{fig:UNet-MedSAM-performance-four-dataset}
\end{figure}

\begin{figure}[h]
\centering
\subfloat[Labeled data from BIDMC]{%
    \label{fig:lb1}%
    \includegraphics[width=0.47\columnwidth]{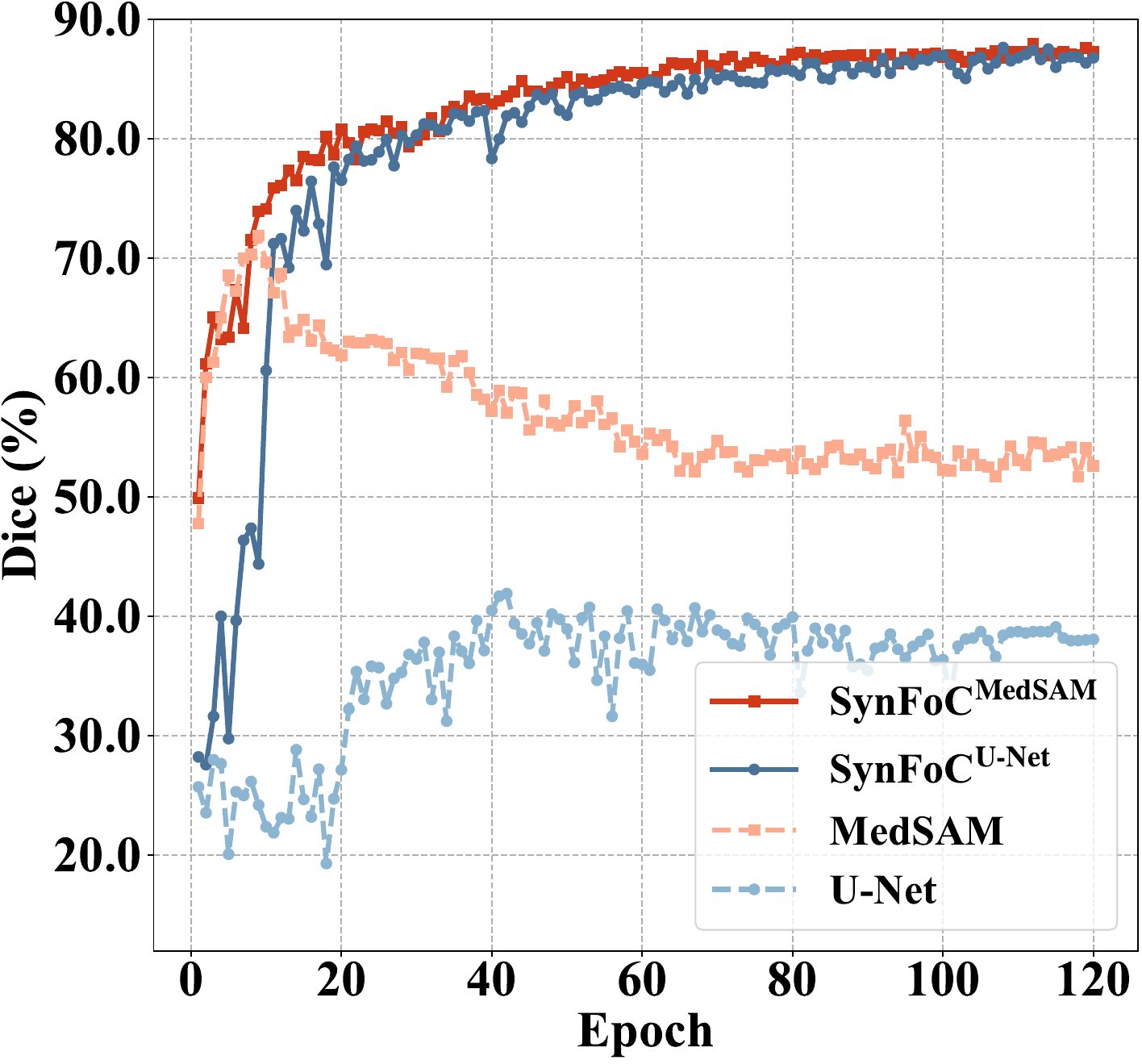}%
}%
\hfill
\subfloat[Labeled data from HK]{%
    \label{fig:lb3}%
    \includegraphics[width=0.47\columnwidth]{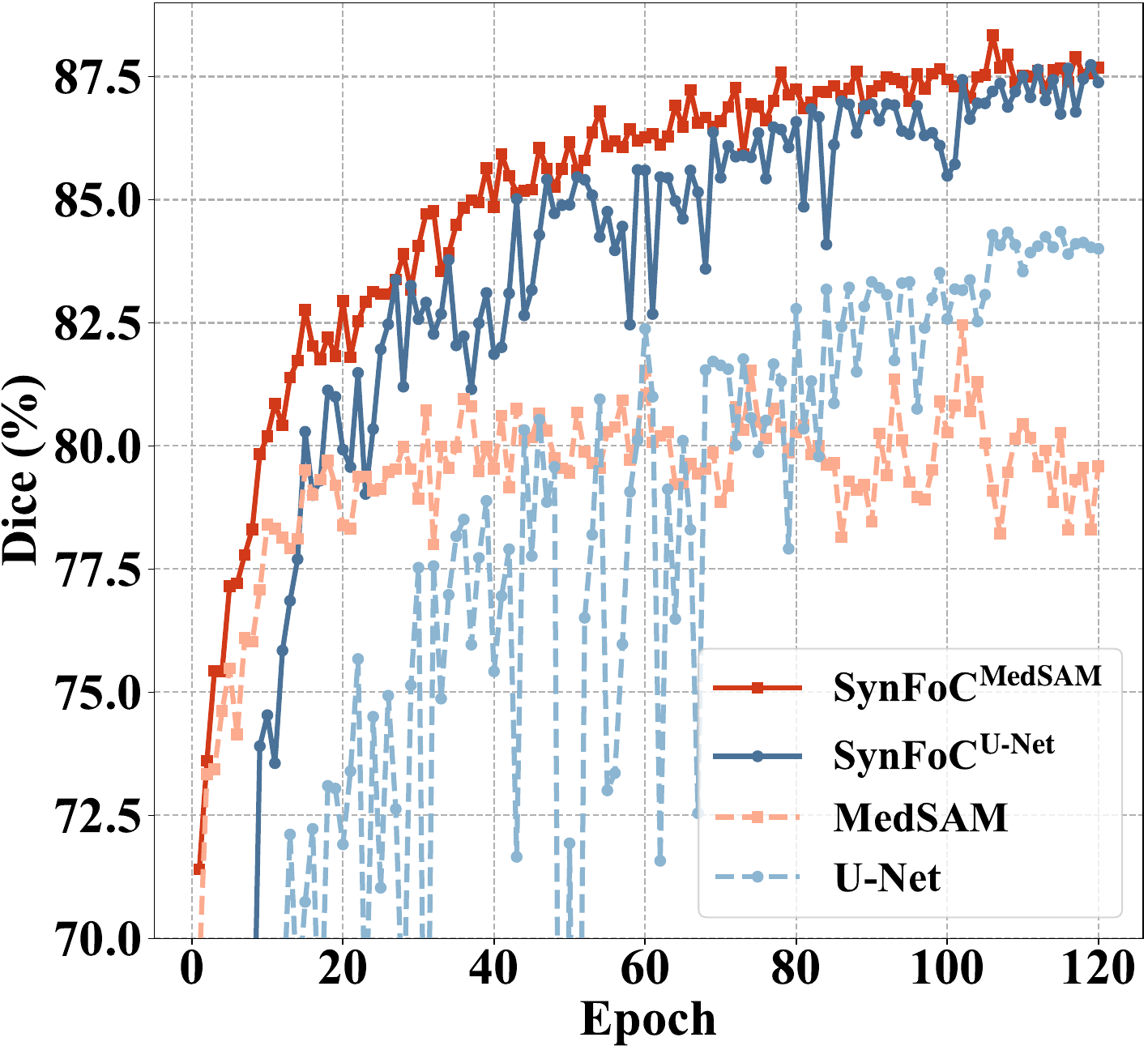}%
}%
\caption{The experimental results on Prostate with 20 labeled data
from BIDMC and HK. Each subplot displays the performance curves of individually trained MedSAM and U-Net , as well as the performance curves of MedSAM and U-Net under SynFoC.}
\label{fig:training_curve}
\vspace{-15pt}
\end{figure}
\textbf{Comparison with Standalone U-Net and MedSAM Across Four Datasets.} We present the advantages of our method over standalone U-Net and MedSAM on four datasets in~\cref{fig:UNet-MedSAM-performance-four-dataset}. Unlike in~\cref{fig:UNet-MedSAM-performance}  (\cref{sec:Synergistic-Training}), where we compare standalone U-Net and MedSAM with SMC-based synergistic training, here we compare them with our overall method, SynFoC. Our method effectively mitigates U-Net's overfitting and further advances MedSAM's performance in downstream tasks across all four datasets, demonstrating superior capabilities in handling significant domain gaps in training data.

In~\cref{fig:training_curve}, we present the performance curves of standalone U-Net and MedSAM, as well as U-Net and MedSAM trained within our SynFoC framework, across two experiments on Prostate dataset (labeled data sourced from BIDMC and HK, respectively). Both U-Net and MedSAM demonstrate significant performance improvements when trained with SynFoC.

\begin{table}[h]
\centering
\footnotesize
\begin{tabular}{cc|cccc}
    \toprule
    U-Net & MedSAM & DSC $\uparrow$ & Jaccard $\uparrow$ & 95HD $\downarrow$ & ASD $\downarrow$\\
    \midrule
    SGD & SGD & 76.56 & 65.85 & 24.51 & 10.70\\
    Adam & Adam & 87.01 & 79.10 & 10.61 & 4.43\\
    Adam & SGD & - & - & - & -\\
    SGD & Adam & \textbf{87.16} & \textbf{79.30} & \textbf{10.26} & \textbf{4.41}\\
    \bottomrule
\end{tabular}
\caption{Ablation study of different optimizer choices.}
\label{tbl:optimizer}
\vspace{-10pt}
\end{table}

\textbf{Different optimizer choices.} We explore the impact of different optimizer choices for U-Net and MedSAM on Prostate dataset. As shown in~\cref{tbl:optimizer}, the best performance is achieved when U-Net and MedSAM are optimized with SGD and Adam, respectively.

\begin{table}[h]
\centering
\footnotesize
\begin{tabular}{c|cccc}
    \toprule
    $\tau$ & DSC $\uparrow$ & Jaccard $\uparrow$ & 95HD $\downarrow$ & ASD $\downarrow$\\
    \midrule
    0.85 & 86.71 & 78.73 & 10.87 & 4.58\\
    0.90 & 87.08 & 79.19 & 10.85 & 4.54\\
    0.95 & \textbf{87.16} & \textbf{79.30} & \textbf{10.26} & \textbf{4.41}\\
    0.98 & 86.68 & 78.78 & 11.55 & 4.98\\
    0.99 & 86.62 & 78.71 & 11.12 & 4.90\\
    \bottomrule
\end{tabular}
\caption{Ablation study of different confidence threshold $\tau$.}
\label{tbl:tau}
\end{table}

\textbf{Discussion on $\tau$.} On Prostate dataset, we investigate the effect of varying the threshold $\tau$ on our method. In~\cref{tbl:tau}, a setting of 0.95 yields the optimal performance, and the results remain stable across other threshold values.

\begin{table*}[t]
\centering
\footnotesize
\begin{tabular}{c|cccccc|cccc}
    \toprule
    Method & \multicolumn{6}{c|}{DSC $\uparrow$} & DSC $\uparrow$ & Jaccard $\uparrow$ & 95HD $\downarrow$ & ASD $\downarrow$\\
    & RUNMC & BMC & HCRUDB & UCL & BIDMC & HK & Avg. & Avg. & Avg. & Avg.\\
    \midrule
    $\text{Standalone}^{\text{U-Net}}$ & 87.74 & 67.96 & 82.45 & 86.31 & 41.91 & 84.35 & 75.12 & 65.76 & 54.67 & 29.08\\
   $\text{Standalone}^{\text{SAM}}$ & 81.15 & 75.26 & 81.93 & 84.78 & 75.09 & 81.26 & 79.91 & 70.35 & 21.59 & 9.22\\
    \color{gray} $\text{SynFoC}^\text{U-Net}$ & \color{gray} 88.05 & \color{gray} 84.42 & \color{gray} 84.11 & \color{gray} 88.00 & \color{gray} 86.84 & \color{gray} 87.31 & \color{gray} 86.46 & \color{gray} 78.84 & \color{gray} 11.66 & \color{gray} 5.03\\
    \rowcolor{Gray} $\text{SynFoC}^\text{SAM}$ & \textbf{88.41} & \textbf{84.94} & \textbf{84.51} & \textbf{88.71} & \textbf{86.71} & \textbf{87.63} & \textbf{86.82} & \textbf{79.21} & \textbf{10.57} & \textbf{4.65}\\
    \bottomrule
\end{tabular}
\caption{Ablation experiments on Prostate dataset.}
\label{tbl:SAM_based_SynFoC}
\end{table*}

\section{The Performance on SSMIS and UDA settings}
\begin{table}[t]
\centering
\footnotesize
\begin{tabular}{c|cc|cc}
    \toprule
    \multirow{2}{*}{Method} & \multicolumn{2}{c|}{Scans used} & \multicolumn{2}{c}{Metrics}\\
    \cmidrule{2-5}
    & L & U & DSC $\uparrow$ & ASD $\downarrow$\\
    \midrule
    \multirow{3}{*}{SupOnly} & 3(5\%) & 0 & 47.83 & 12.62\\
    & 7(\%10) & 0 & 79.41 & 2.70\\
    & 70(All) & 0 & 91.44 & 0.99\\
    \midrule
    SS-Net~\cite{wu2022exploring} & \multirow{4}{*}{3(5\%)} & \multirow{4}{*}{67(95\%)} & 65.83 & 2.28\\
    BCP~\cite{bai2023bidirectional} & & & 87.59 & \underline{0.67}\\
    ABD~\cite{chi2024adaptive} & & & \textbf{88.96} & \textbf{0.52}\\
    SynFoC & & & \underline{88.32} & 0.70\\
    \midrule
    SS-Net~\cite{wu2022exploring} & \multirow{4}{*}{7(10\%)} & \multirow{4}{*}{63(90\%)} & 86.78 & 1.40\\
    BCP~\cite{bai2023bidirectional} & & & 88.84 & \underline{1.17}\\
    ABD~\cite{chi2024adaptive} & & & \textbf{89.81} & \textbf{0.49}\\
    SynFoC & & & \underline{89.68} & \underline{1.17}\\
    \bottomrule
\end{tabular}
\caption{Comparison of different methods on ACDC dataset.}
\label{SSMIS}
\end{table}


\begin{table}[t]
\centering
\footnotesize
\begin{tabular}{c|cccc}
    \toprule
    \multirow{2}{*}{Method} & \multicolumn{4}{c}{DSC $\uparrow$}\\
    & MYO & LV & RV & Avg.\\
    \midrule
    \multicolumn{5}{c}{Adaptation from \textbf{CT} to \textbf{MRI}}\\
    \midrule
    NoAdapt & 14.50 & 34.51 & 31.10 & 26.70\\
    SIFA~\cite{chen2020unsupervised} & 67.69 & 83.31 & \textbf{79.04} & 76.68\\
    UDA-VAE~\cite{wu2021unsupervised} & 68.42 & 84.41 & 72.59 & 75.14\\
    UDA-VAE++~\cite{lu2022unsupervised} & \underline{70.75} & \textbf{88.64} & 75.82 & \underline{78.40}\\
    SynFoC & \textbf{71.47} & \underline{86.90} & \underline{78.81} & \textbf{79.06}\\
    \midrule
    \multicolumn{5}{c}{Adaptation from \textbf{MRI} to \textbf{CT}}\\
    \midrule
    NoAdapt & 12.32 & 30.24 & 37.25 & 26.60\\
    SIFA~\cite{chen2020unsupervised} & 60.89 & 79.32 & \underline{82.39} & 74.20\\
    UDA-VAE~\cite{wu2021unsupervised} & 58.58 & 79.43 & 80.43 & 72.81\\
    UDA-VAE++~\cite{lu2022unsupervised} & \underline{68.74} & \underline{85.08} & 81.42 & \underline{78.41}\\
    SynFoC & \textbf{78.26} & \textbf{88.25} & \textbf{82.97} & \textbf{83.16}\\
    \bottomrule
\end{tabular}
\caption{Comparison of different methods on MSCMRSeg dataset.}
\label{UDA}
\end{table}

Our SynFoC offers a general solution to address domain shifts and limited labeled data. As shown in~\cref{SSMIS,UDA}, we conduct experiments on the ACDC dataset~\cite{bernard2018deep} (containing 100 patients' scans) and the MSCMRSeg dataset~\cite{zhuang2018multivariate} (containing 35 labeled CT images and 45 labeled LGE-MRI images), demonstrating its competitive performance in traditional SSMIS and UDA settings.

\section{Common Challenges of Foundation Models}
As shown in~\cref{tbl:SAM_based_SynFoC}, SAM also struggles to correct high-confidence mispredictions. Due to large-scale pretraining, the issue of error accumulation is a common challenge for foundation models in downstream tasks. Our SynFoC method is not limited to the combination of U-Net and MedSAM. Through the Synergistic training of conventional and foundation models, we achieve significant performance improvements for both models.

\section{Limitations and Future Works}
\begin{figure}[h]
    \centering
    \includegraphics[width=0.9\linewidth]{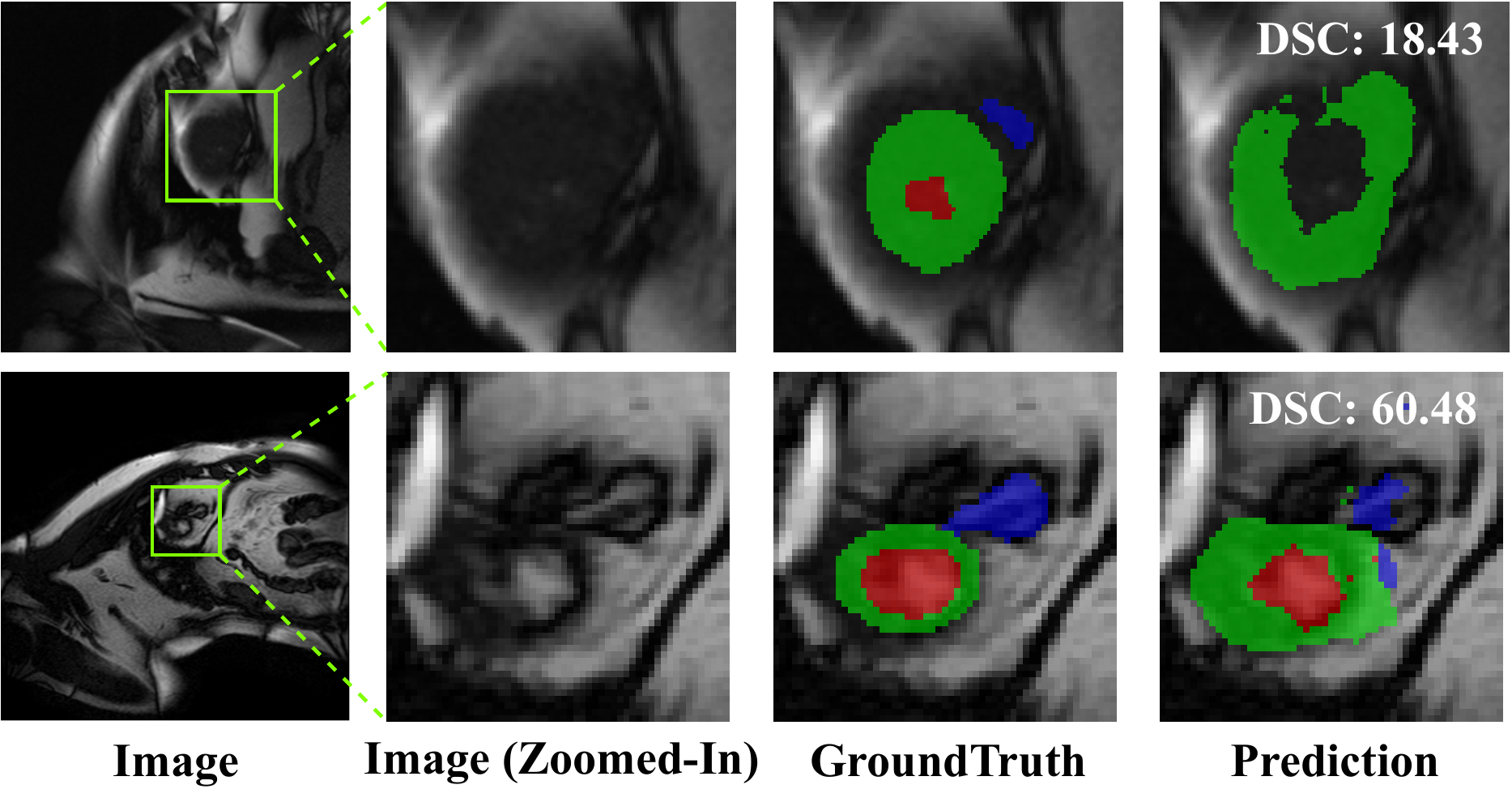}
\caption{Visual results of error cases on M\&Ms dataset.}
\label{fig:error_cases}
\end{figure}

As shown in~\cref{fig:error_cases}, in the experiments on M\&Ms dataset, by deeply analyzing the results, we found that SynFoC and most existing methods struggle with extremely small targets. Visual analysis of error cases reveals that tiny size and low boundary contrast often lead to over- or under-segmentation. Additionally, our method focuses on 2D medical image segmentation and lacks exploration in 3D medical image segmentation. Future work could enhance the precise segmentation of extremely small targets and extend the framework to 3D medical images.

{
    \clearpage
    \small
    \bibliographystyle{ieeenat_fullname}
    \bibliography{main}
}


\end{document}